\documentclass{article}

\usepackage[nonatbib, main, final]{neurips_2025}

\usepackage[utf8]{inputenc} % allow utf-8 input
\usepackage[T1]{fontenc}    % use 8-bit T1 fonts
\usepackage[hidelinks]{hyperref}
\usepackage{url}            % simple URL typesetting
\usepackage{booktabs}       % professional-quality tables
\usepackage{amsfonts}       % blackboard math symbols
\usepackage{nicefrac}       % compact symbols for 1/2, etc.
\usepackage{microtype}      % microtypography
\usepackage{xcolor}         % colors

\usepackage{multirow}
\usepackage{graphicx}
\usepackage{colortbl}
\usepackage[table]{xcolor}
\usepackage{wrapfig}
\usepackage{caption}
\usepackage{amsmath}
\usepackage{enumitem}

\usepackage[scaled=0.8]{DejaVuSans}

\title{\begin{minipage}{0.1\textwidth}% adapt widths of minipages to your needs
\includegraphics[width=\linewidth]{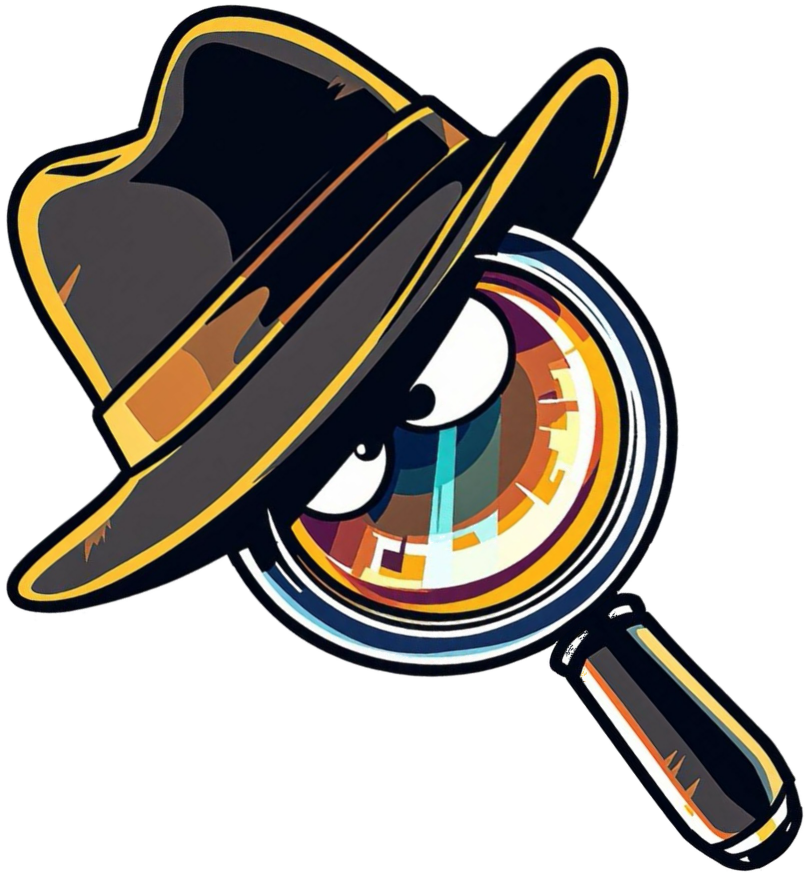}
\end{minipage}%
\hfill%
\begin{minipage}{0.88\textwidth}\centering
ForgerySleuth: Empowering Multimodal Large Language Models for Image Manipulation Detection
\end{minipage}}

\author{
Zhihao Sun$^{1,2}$, Haoran Jiang$^{1,2}$, Haoran Chen$^{1,2}$, Yixin Cao$^{1,2}$, Xipeng Qiu$^{1,2}$, \\
\textbf{Zuxuan Wu}$^{1,2\dag}$, \textbf{Yu-Gang Jiang}$^{1,2}$\\[6pt]
$^{1}$Shanghai Key Lab of Intell. Info. Processing, School of CS, Fudan University\\
$^{2}$Shanghai Collaborative Innovation Center of Intelligent Visual Computing\\[6pt]
\textcolor{magenta}{\href{https://github.com/sunzhihao18/ForgerySleuth}{https://github.com/sunzhihao18/ForgerySleuth}}
}

\begin{document}

\footnotetext{$\dag$: corresponding author.}

\maketitle

\begin{abstract}

Multimodal large language models have unlocked new possibilities for various multimodal tasks. However, their potential in image manipulation detection remains unexplored. When directly applied to the IMD task, M-LLMs often produce reasoning texts that suffer from hallucinations and overthinking. To address this, we propose ForgerySleuth, which leverages M-LLMs to perform comprehensive clue fusion and generate segmentation outputs indicating specific regions that are tampered with. Moreover, we construct the ForgeryAnalysis dataset through the Chain-of-Clues prompt, which includes analysis and reasoning text to upgrade the image manipulation detection task. A data engine is also introduced to build a larger-scale dataset for the pre-training phase. Our extensive experiments demonstrate the effectiveness of ForgeryAnalysis and show that ForgerySleuth significantly outperforms existing methods in generalization, robustness, and explainability.

\end{abstract}    
\section{Introduction}
\label{sec:intro}

\begin{figure}[h]
  \centering
  \includegraphics[width=\linewidth]{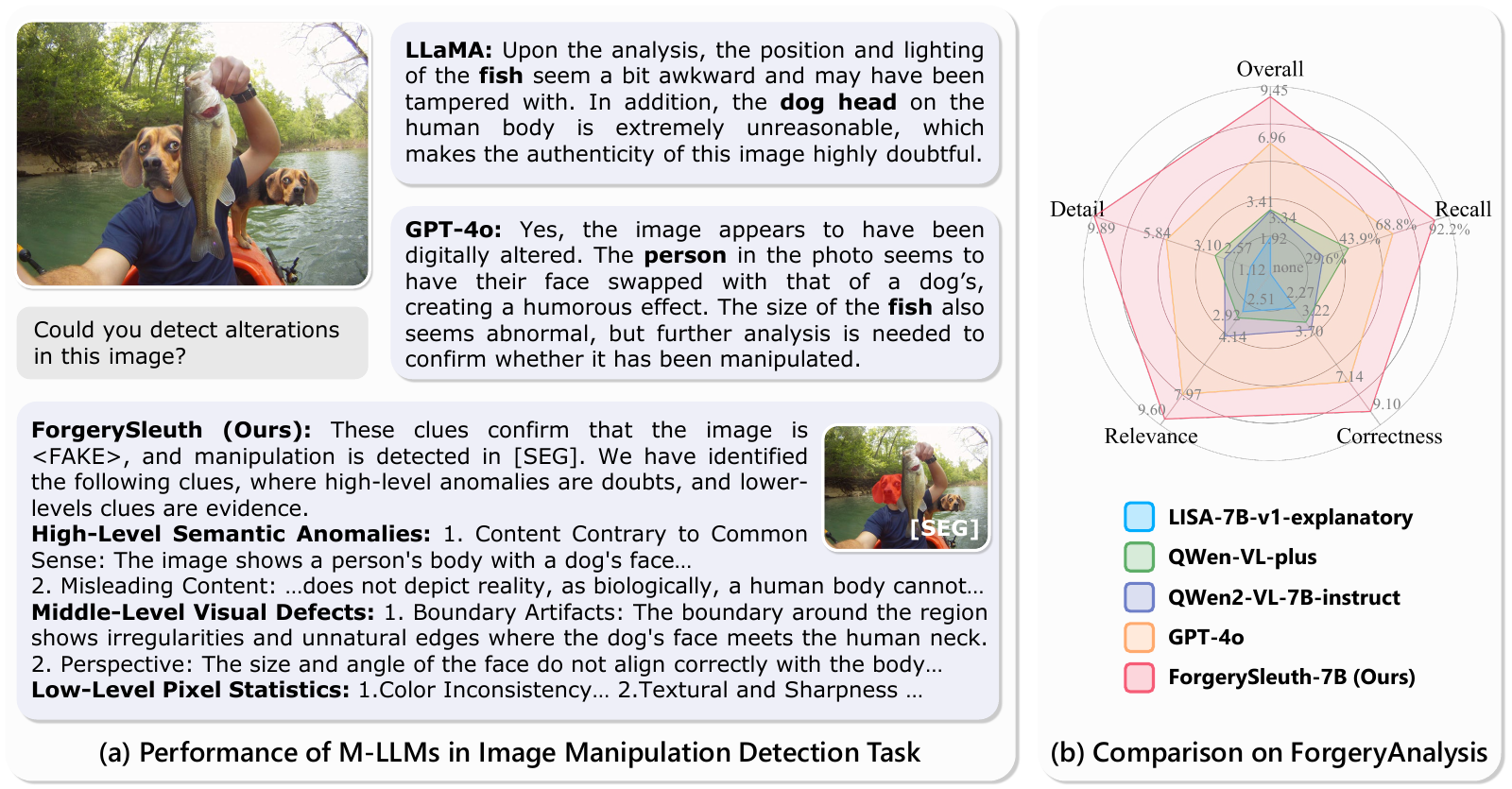}
  \caption{Performance and comparison of existing M-LLMs on the image manipulation detection task. Our ForgerySleuth assistant provides explanatory analysis with Chain-of-Clues and demonstrates the best forgery analysis capabilities.}
  \label{figure:teaser}
\end{figure}

Recent advancements in multimodal large language models (M-LLMs) \cite{yin2023survey} have unlocked new possibilities across a variety of tasks, propelling advancements in numerous traditional research fields~\cite{Med-flamingo, mplug-paperowl, cui2024survey}. Despite these advancements, few attempts have been made to enhance the ability of multimodal large language models on the image manipulation detection task (IMD). In fact, with the rise of advanced image generation and editing tools, both creative and malicious alterations to visual content have become increasingly difficult to detect. As a result, identifying manipulated images has become essential for maintaining the integrity of digital media \cite{schetinger2017humans, nightingale2017can, lu2024seeing}. 

While appealing, directly applying M-LLMs for IMD produces results in text, which falls short even compared with traditional IMD methods that are able to generate segmentation masks to highlight tampered regions. In addition, M-LLMs are often prone to hallucinations and overthinking, making them unreliable for precise manipulation detection. For instance, as shown in Figure \ref{figure:teaser} (a), M-LLMs may incorrectly identify non-tampered regions, such as the ``fish'' area, as altered. This highlights the need to enhance the reasoning capabilities of M-LLMs for tampered region detection. Furthermore, M-LLMs, pre-trained on vast datasets, excel in recognizing authentic image patterns and world knowledge, yet lack the ability to combine low-level statistical features to pinpoint manipulation evidence, which has proven crucial for IMD \cite{Mantra-Net, SPAN, MVSS-Net, SAFL-Net}.

In this paper, we explore how to unleash the power of M-LLMs for IMD tasks. We introduce carefully designed modules that empower an M-LLM as an IMD expert, named ForgerySleuth. ForgerySleuth is expected to provide a textual explanation of detected clues with the reasoning process, along with a segmentation mask to highlight tampered regions. Inspired by the fact that existing IMD methods rely on low-level features, such as noise patterns, to localize precise tampered regions, we aim to capture similar low-level features through ForgerySleuth. To do so, ForgerySleuth integrates M-LLMs with a trace encoder, enabling the model to leverage world knowledge to detect high-level semantic anomalies while still capturing low-level forgery traces. Additionally, inspired by LISA \cite{LISA}, we introduce a vision decoder with a fusion mechanism that uses attention to combine high-level anomalies in the LLM tokens and low-level traces in the trace embedding, ensuring the generation of accurate segmentation masks.

To further enhance the reasoning capability of M-LLMs on manipulation detection, we propose a supervised fine-tuning (SFT) dataset ForgeryAnalysis specifically tailored for the IMD task. Each entry in ForgeryAnalysis is initially generated by GPT-4o using a novel Chain-of-Clues prompt. Specifically, we ask GPT-4o to produce a detailed thought process and provide reasoning for why a particular region is tampered with, including high-level semantic anomalies (\textit{e.g.}, content that contradicts common sense), mid-level visual defects (such as lighting inconsistencies), and low-level pixel information (such as color, texture, \textit{etc}.). The generated entries are then reviewed and refined by experts. Additionally, we build a data engine based on this dataset to automate forgery analysis, enabling us to create a larger-scale ForgeryAnalysis-PT dataset for pre-training. 

Extensive experiments on popular benchmarks demonstrate the success of ForgerySleuth in pixel-level manipulation localization and text-based forgery analysis. Specifically, our approach outperforms the current SoTA method by up to 24.7\% in pixel-level localization tasks. Moreover, in the ForgeryAnalysis-Eval comprehensive scoring, our method surpasses the best available model, GPT-4o, achieving an improvement of 35.8\%. In summary, our main contributions include:

\begin{itemize}[leftmargin=*]
    \item \textbf{Novel Exploration.} We explored the role of M-LLMs in image manipulation detection, upgrading the manipulation detection task by incorporating clues analysis and reasoning.
    \item \textbf{Valuable Dataset.} We constructed ForgeryAnalysis dataset, providing instructions for analysis and reasoning through Chain-of-Clues prompting. Additionally, we developed a data engine to automate forgery analysis, enabling the creation of a large-scale dataset.
    \item \textbf{Practical Framework.} We introduced ForgerySleuth assistant framework, which integrates M-LLMs with a trace encoder to leverage multi-level clues. The vision decoder with a fusion mechanism enables comprehensive clues fusion and segmentation outputs.
\end{itemize}

\section{Related Work}
\label{sec:related}

\subsection{LLMs and Multimodal LLMs}

The success of large language models in various natural language processing tasks has led researchers to explore their integration with vision modalities, resulting in the development of M-LLMs. BLIP-2 \cite{BLIP-2} introduces a visual encoder to process image features. LLaVA \cite{LLaVA} aligns image and text features to achieve comprehensive visual and language understanding. Researchers also utilize prompt engineering to connect independent vision and language modules via API calls without end-to-end training~\cite{HuggingGPT, VisualChatGPT, GPT4tools}. However, while these approaches enable M-LLMs to perceive, the intersection with vision-centric tasks, such as segmentation, remains underexplored. Additionally, VisionLLM \cite{VisionLLM} and LISA \cite{LISA} effectively integrate segmentation capabilities into M-LLMs, making them support vision-centric tasks, such as segmentation.

With advancements in fundamental reasoning and multimodal information processing, M-LLMs have demonstrated impressive proficiency across a diverse range of tasks, including image captioning and video understanding \cite{yin2023survey}. Moreover, M-LLMs have been developed to address more complex real-world tasks in robotics, such as embodied agents \cite{LEO, Kosmos-2} and autonomous driving \cite{cui2024survey}. However, integrating M-LLMs into the field of image manipulation detection remains unexplored. While M-LLMs possess valuable world knowledge and can potentially detect high-level anomalies, M-LLMs are often prone to hallucinations and overthinking, making them unreliable for precise manipulation detection. Furthermore, there is no existing IMD dataset with analysis instructions for supervised fine-tuning, which further restricts their capabilities.

% \zx{not clear}.  they lack the capability of modeling low-level features like noise patterns~\cite{Mantra-Net, SPAN, SAFL-Net}, which are crucial for manipulation detection

\subsection{Image Manipulation Detection}

Image manipulation detection is a critical task in digital image forensics. The task has evolved beyond merely determining whether an image is authentic \cite{chen2015median, bayar2016deep}; it involves localizing tampered regions and providing segmentation masks \cite{MFCN, bappy2017detection}, which leads to more intuitive results. 
Early attempts \cite{CFA, mahdian2009using, johnson2005exposing, liu2011identifying, wengrowski2017reflection} identified anomalies and designed corresponding hand-engineered features. These efforts systematically use various tampering clues, laying a solid foundation for the field. However, such hand-engineered features are specific to certain tampering types, which limits their applicability in real-world scenarios.

Recent approaches have shifted to a more general capability of identifying complex and unknown manipulations. Semantic-agnostic features, less dependent on specific content, are thought to provide better generalization \cite{MVSS-Net, SAFL-Net}. Common strategies include incorporating filters or extractors to capture low-level noise features \cite{Mantra-Net, SPAN, TruFor, UnionFormer} and high-frequency features \cite{ObjectFormer}, and using content features extracted from the image view as a supplement to detect manipulation traces \cite{MVSS-Net, TruFor, UnionFormer}. However, many of these features are learned implicitly by the network, which limits their explainability.
Studies also \cite{SAFL-Net, ObjectFormer} detect anomalies by comparing patch-level or object-level features. However, capturing high-level semantic anomalies, such as content that conflicts with common sense or physical laws, is still challenging.
In this work, we extend the task by presenting a reasoning process with multiple levels of clues expressed in natural language, making the detection results more comprehensible. Our proposed ForgerySleuth framework leverages M-LLMs to address this challenge, effectively leveraging world knowledge to detect high-level semantic anomalies while still capturing low-level forgery traces using a trace encoder.
FakeShield \cite{FakeShield} is a concurrent work that proposes a multimodal large model for image manipulation detection.

\section{ForgeryAnalysis Dataset}
\label{sec:dataset}

\begin{figure*}[t]
  \centering
  \includegraphics[width=\linewidth]{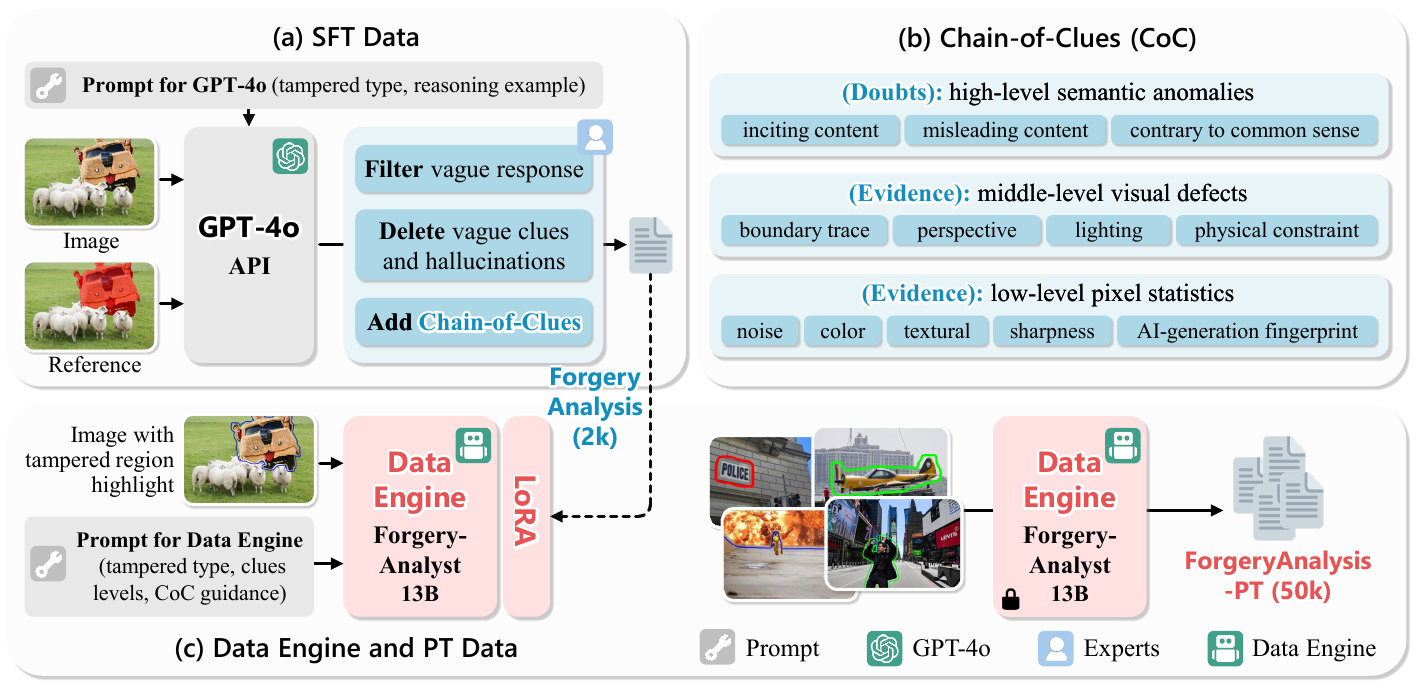}
  \caption{\textbf{ForgeryAnalysis Dataset Construction Pipeline.} Our pipeline begins with (a) GPT-4o generating initial analyses for manipulated images with annotated regions, followed by human expert review. The refined analyses are organized into (b) the Chain-of-Clues format. This human-curated ForgeryAnalysis (2k) dataset is used to train a data engine. Finally, (c) this data engine generates ForgeryAnalysis-PT, a larger-scale dataset for model pre-training.}
  \label{figure:dataset}
  \vspace{-1em}
\end{figure*}

Our goal is to leverage existing M-LLMs to construct a high-quality dataset for IMD. We first describe how M-LLMs are utilized to generate initial clue analyses from various types of manipulated image sources in Section \ref{subsec:sft_data}. These analyses are then meticulously refined by experts to create a high-quality dataset of 2,370 samples, which are used for the supervised fine-tuning (SFT) phase and the evaluation of M-LLMs. Section \ref{subsec:pt_data} describes our proposed data engine, which expands the 2k high-quality analysis instructions from Section \ref{subsec:sft_data} to 50k. This expansion supports the pre-training phase while maintaining quality standards. Detailed statistics of our dataset, along with examples of tampering analysis instructions, are provided in Section \ref{sec:appendices_dataset}. 

%By addressing a gap in the IMD field—where instruction-following data was previously unavailable—we offer a diverse range of image manipulation types, ensure quality through advanced models, and provide scalability through the data engine.

\subsection{ForgeryAnalysis Data}
\label{subsec:sft_data}

Due to the absence of instruction-following datasets that support clue analysis, we introduce the ForgeryAnalysis dataset to support the supervised fine-tuning (SFT) and evaluation of M-LLMs. As illustrated in Figure \ref{figure:dataset} (a), for each image selected from various data sources, we utilize the advanced M-LLM, GPT-4o, to generate initial clue analyses. Experts then carefully revise these analyses to eliminate hallucinations and ensure high quality. Furthermore, we have integrated a Chain-of-Clues structure to enhance reasoning capabilities.

\vspace{0.05in}
\noindent\textbf{Data Source.} 
To ensure a diverse dataset that covers various types of manipulation, we collect 4,000 tampered images from existing IMD datasets, including MIML \cite{MIML}, CASIA2 \cite{CASIA}, DEFACTO \cite{DEFACTO}, and AutoSplice \cite{AutoSplice}.  Each image is accompanied by annotations indicating the tampered regions. These sources include common real-world manipulation techniques such as splicing, copy-move, object removal, AI generation, and Photoshop edits.

\vspace{0.05in}
\noindent\textbf{Instruction Design.} We aim to provide highly specific and content-driven descriptions of clues for each image, rather than vague or general analysis. We refer to the clues and evidence widely used in digital image forensics and manipulation detection \cite{wang2009survey, zheng2019survey} to design a more reliable and effective instruction structure. The key clues include but are not limited to 1) low-level pixel statistics (\textit{e.g.}, noise, color, texture, sharpness, and AI-generation fingerprints), 2) middle-level visual defects (\textit{e.g.}, traces of tampered boundaries, lighting inconsistencies, and perspective relationships), and 3) high-level semantic anomalies (\textit{e.g.}, content that contradicts common sense, incites, or misleads).

The prompt instructs GPT-4o to assume it has detected manipulation in the highlighted region of the reference image and to analyze its detection based on clues from various levels and aspects. We provide the tampering type to help GPT-4o focus on relevant clues, along with a reasoning example to guide the content and format of the output. We also instruct GPT-4o to incorporate corresponding world knowledge, such as well-known individuals or landmarks.

\vspace{0.02in}
\noindent\textbf{Chain-of-Clues.}
The responses generated by GPT-4o are subsequently revised by experts to ensure quality. Responses lacking detailed content analysis are filtered out, retaining only high-quality responses as draft data. The experts carefully review vague and incorrect statements that may arise from hallucinations, removing irrelevant clues and evidence. Inspired by works such as \cite{CoT, ToT} that introduce Chain-of-Thought (CoT) prompting and demonstrate its effectiveness in enhancing the step-by-step reasoning capabilities of LLMs, we propose a Chain-of-Clues (CoC) prompting approach, illustrated in Figure \ref{figure:dataset} (b). The reasoning process begins with ``unveiling doubts'' using high-level clues, followed by ``pinpointing evidence'' through middle-level and low-level features. We organize the clues according to this structure, creating a coherent chain of clues, which results in 2,370 manually revised samples. To ensure accuracy, we conduct additional cross-validation with more than two experts, selecting 618 samples for evaluation. The remaining data are used for supervised fine-tuning.

\subsection{ForgeryAnalyst Engine}
\label{subsec:pt_data}
The pre-training phase requires a larger-scale dataset compared to the supervised fine-tuning phase. However, the high cost of GPT-4o and the need for expert revisions make expanding the dataset a challenge. To overcome this limitation and create a pre-training dataset, we introduce an additional data engine \cite{SAM, SAM2, Depth-Anything, ShareGPT4v} that scales up the training set. Specifically, we fine-tune LLaVA-v1.5-13B using LoRA \cite{LoRA} on the ForgeryAnalysis-SFT dataset, aiming to replicate the ability of GPT-4o to analyze clues while having experts eliminate any hallucinations. The resulting model is used as our data engine that is able to annotate data automatically. In particular, we select 50k images from existing public datasets. The data engine receives input that includes explicit information about the tampered region, aiming to generate more precise and comprehensive clue analyses. It outputs manipulation analyses organized in the CoC format, as illustrated in Figure \ref{figure:dataset} (c). For more detailed information on the data engine, including the specific prompts used, please refer to Section \ref{sec:appendices_engine}, and Figure \ref{figure:sup_fa_pt} for output examples. The entire analysis generation process takes approximately 16 A800 GPU days. We refer to this dataset as ForgeryAnalysis-PT.

% 28s/it -> 16 A800 GPU days/50k data

\section{ForgerySleuth}
\label{sec:method}

% \begin{wrapfigure}{r}{0.5\textwidth}
\begin{figure}[t]
    \centering
    \includegraphics[width=\textwidth]{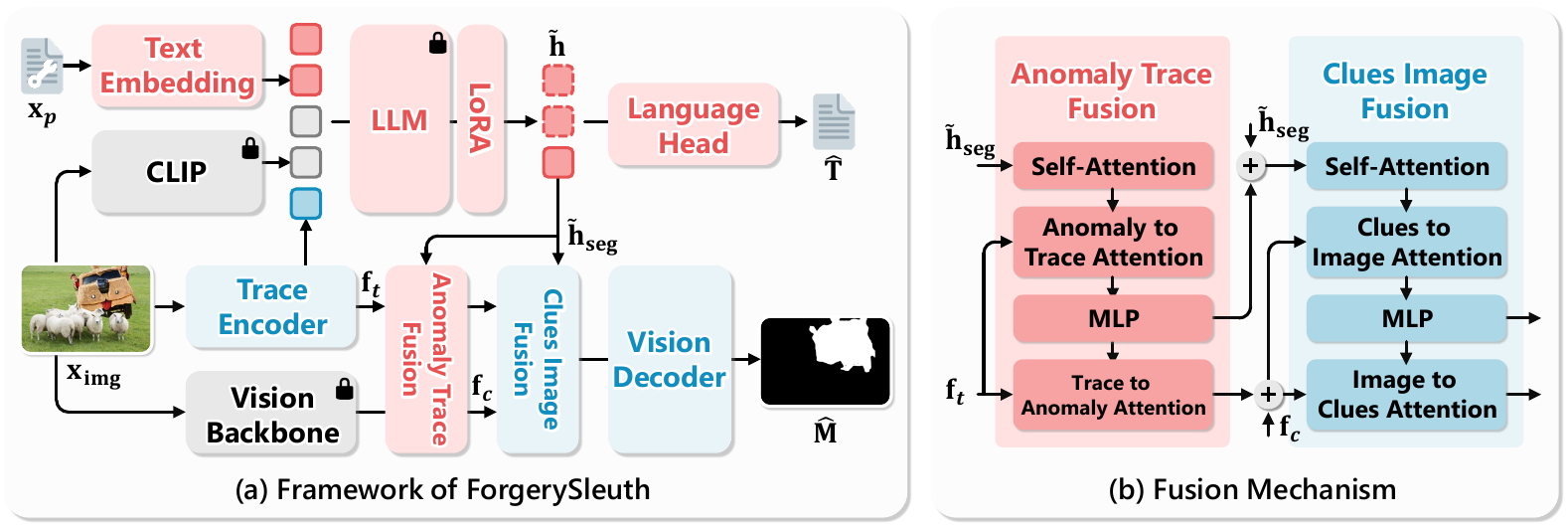}
    \caption{\textbf{Framework of ForgerySleuth.} Given an image $\mathbf{x}_\text{img}$ and a prompt query $\mathbf{x}_{p}$, the M-LLM $\mathcal{F}_{m}$ detects high-level semantic anomalies and generates a textual output $\hat{\mathbf{T}}$. The trace encoder $\mathcal{F}_{t}$ captures low-level, semantic-agnostic features. The vision decoder $\mathcal{D}$ fuses vision embeddings with the prompt embedding corresponding to \texttt{[SEG]} token to generate the segmentation mask $\hat{\mathbf{M}}$. LoRA is utilized in trainable modules for fine-tuning.}
    \label{figure:framework}
\end{figure}
% \end{wrapfigure}

Given an input image for detection and a prompt specifying the detection request, our goal is to output a binary segmentation mask $\hat{\mathbf{M}}$ of the tampered regions, as well as text $\hat{\mathbf{T}}$ that includes reasoning and evidence. This poses a challenge as the model needs to bridge vision and language modalities and capture tampering features across multiple levels. 
 To address this, ForgerySleuth uses M-LLMs to connect vision and language modalities, enabling the description of detected clues and evidence in textual form. Additionally, we incorporate a vision decoder into the multimodal large language models to perform dense prediction, generating the tampered region mask for this specific task. The pipeline of our framework is illustrated in Figure \ref{figure:framework} (a).

More formally, we extend the original LLM vocabulary with new tokens, including \texttt{[SEG]}, which requests segmentation output, and \texttt{<REAL>} and \texttt{<FAKE>}, which indicate the image classification results. Given the input image $\mathbf{x}_\text{img}$ to be detected along with the prompt $\mathbf{x}_{p}$, we first feed them into the M-LLM $\mathcal{F}_{m}$, which outputs the hidden embedding $\tilde{h}$ at the last layer of LLM. We then extract the embedding $\mathbf{\tilde{h}}_\text{seg}$ corresponding to the \texttt{[SEG]} token from the hidden embedding $\mathbf{\tilde{h}}$ and apply an MLP projection layer $\gamma$ to obtain $\mathbf{h}_\text{seg}$. 
The process can be formulated as
\begin{equation}
    \mathbf{\tilde{h}} = \mathcal{F}_{m}^{\prime}(\mathbf{x}_\text{img}, \mathbf{x}_{p}),
\end{equation}
\begin{equation}
    \mathbf{h}_\text{seg} = \gamma(\mathbf{\tilde{h}}_\text{seg}).
\end{equation}
$\mathcal{F}_{m}^{\prime}$ denotes M-LLM without the language head. This embedding represents the suspicious anomalous regions the M-LLM detects and serves as the input to guide the subsequent vision decoder.

Inspired by existing IMD methods, we propose an independent trace encoder $\mathcal{F}_{t}$ to focus on low-level features, which complements the relatively high-level vision and semantic features discovered by the M-LLM and provides more reliable tampering evidence, helping to minimize hallucinations from the LLM. Specifically, constrained convolutions \cite{Bayar} are employed with residual connections \cite{ResNet} at the front part of the encoder to suppress the image content and learn manipulation features adaptively. The input image $\mathbf{x}_\text{img}$ is also fed into encoder $\mathcal{F}_{t}$, producing dense manipulation trace features $\mathbf{f}_{t}$, which can be expressed by
\begin{equation}
    \mathbf{f}_{t} = \mathcal{F}_{t}(\mathbf{x}_\text{img}).
\end{equation}

The vision backbone $\mathcal{F}_{v}$ simultaneously extracts dense vision content feature $\mathbf{f}_{c}$ to support more precise segmentation. Finally, $\mathbf{f}_{c}$, $\mathbf{f}_{t}$ and $\mathbf{h}_\text{seg}$ are fed to the vision decoder $\mathcal{D}$ to generate the segmentation mask $\mathbf{\hat{M}}$ which indicates the tampered region. The language modeling head  $\mathcal{H}$ processes $\mathbf{\tilde{h}}$ and outputs $\mathbf{\hat{y}}_\text{txt}$, containing an analysis of the reasoning and evidence. It can be formulated as
\begin{equation}
    \mathbf{f}_{c} = \mathcal{F}_{v}(\mathbf{x}_\text{img}),
\end{equation}
\begin{equation}
    \mathbf{\hat{M}} = \mathcal{D}(\mathbf{f}_{c},\mathbf{f}_{t},\mathbf{h}_\text{seg}), \quad
 \mathbf{\hat{T}} = \mathcal{H}(\mathbf{\tilde{h}}).
\end{equation}

\vspace{0.05in}
\noindent\textbf{Fusion Mechanism.} To integrate the image content embedding $\mathbf{f}_{c}$, trace embedding $\mathbf{f}_{t}$, and the LLM output tokens $\mathbf{h}_\text{seg}$ obtained from the M-LLM, we take inspiration from Transformer segmentation models \cite{DETR, MaskFormer, SAM} and design a vision decoder with a fusion attention mechanism, as illustrated in Figure \ref{figure:framework} (b). Here, the first layer of the module computes attention between anomalies in the LLM output tokens and traces in the trace embeddings, facilitating the organization and pinpointing of clues. The subsequent layers focus on attention between refined clues in the upgraded tokens and content in the image embeddings, enabling more precise segmentation of the tampered regions.

\vspace{0.05in}
\noindent\textbf{Learning Objective.} Our framework is trained end-to-end using both the reasoning text loss $\mathcal{L}_\text{txt}$ and the tampered region mask loss $\mathcal{L}_\text{mask}$. The final learning objective $\mathcal{L}$ is formulated as a weighted sum of these losses, with weight $\lambda_\text{txt}$ and $\lambda_\text{mask}$, as
\begin{equation}
    \mathcal{L} = \lambda_\text{txt}\mathcal{L}_\text{txt} + \lambda_\text{mask}\mathcal{L}_\text{mask}.
\end{equation}

Specifically, $\mathcal{L}_\text{txt}$ is the auto-regressive cross-entropy (CE) loss for reasoning text generation, guiding the model to collect multi-level clues, while $\mathcal{L}_\text{mask}$ is the mask loss, which encourages the model to produce precise segmentation results. $\mathcal{L}_\text{mask}$ is implemented by the combination of binary cross-entropy (BCE) loss and DICE loss, with respective weights $\lambda_{bce}$ and $\lambda_\text{dice}$. Given the ground-truth targets $\mathbf{T}$ and $\mathbf{M}$, the final loss functions are defined as
\begin{equation}
    \mathcal{L}_\text{txt} = \textbf{CE}(\mathbf{\hat{T}}, \mathbf{T}),
\end{equation}
\begin{equation}
    \mathcal{L}_\text{mask} = \lambda_\text{bce}\textbf{BCE}(\mathbf{\hat{M}}, \mathbf{M}) + \lambda_\text{dice}\textbf{DICE}(\mathbf{\hat{M}}, \mathbf{M}).
\end{equation}
It is noteworthy that no separate loss is required for the classification result, as the supervision for \texttt{<REAL>} and \texttt{<FAKE>} tokens are integrated into the text loss $\mathcal{L}_\text{txt}$.

\subsection{Training Strategy}

\vspace{0.05in}
\noindent\textbf{Trainable Parameter.} To preserve the world knowledge and the normal patterns of authentic images learned by the pre-trained M-LLM $\mathcal{F}_{m}$, we adopt LoRA \cite{LoRA} for efficient fine-tuning. The vision backbone $\mathcal{F}_{v}$ is entirely frozen to retain the capacity for modeling image content features, which are crucial for accurate segmentation. Both the trace encoder $\mathcal{F}_{t}$ and the vision decoder $\mathcal{D}$ are fully trainable and fine-tuned. Additionally, the token embeddings of the LLM, the language modeling head $\mathcal{H}$, and the projection layer $\gamma$ are updated during training. Despite the large model scale, only 5.47\% of the parameters are trainable, making end-to-end training more efficient.

\vspace{0.05in}
\noindent\textbf{Data Formulation.} The training process involves two phases: \textbf{1) Pre-training Phase}: (a) In the first stage, we aim to align the framework modules and ensure that the model can perform basic segmentation and visual reasoning tasks. For foundational segmentation abilities, we use semantic segmentation datasets such as ADE20k \cite{ADE20k} and COCO-Stuff \cite{COCO-Stuff}, transforming these datasets into visual question-answer pairs using class names as questions. Additionally, we incorporate ReasonSeg \cite{LISA} to strengthen visual reasoning capabilities. (b) The ForgeryAnalysis-PT dataset focuses on the IMD task, including analysis instructions that enable the model to recognize tampering traces and identify clues. We also utilize public IMD datasets, including MIML \cite{MIML} and CASIA2 \cite{CASIA}, with prompts randomly selected from simple or vague responses. \textbf{2) Supervised Fine-tuning Phase}: The ForgeryAnalysis-SFT dataset, meticulously revised by experts to ensure the accuracy of reasoning and analysis, is used for final supervised fine-tuning to standardize the analysis and output.

\section{Experiment}
\label{sec:experiment}

\vspace{-0.5em}

\subsection{Experimental Setting}

\begin{table*}[t]
    \centering
    \caption{Manipulation localization results comparing ForgerySleuth with SoTA methods.}
    % for optimal and fixed thresholds (0.5)
    \resizebox{\linewidth}{!}{
        \setlength{\tabcolsep}{1mm}{
            \begin{tabular}{lcccccccccc}
                \toprule
                \multirow{2}[1]{*}{\textbf{Method}} & \multicolumn{5}{c}{\textbf{Optimal Threshold F1}} & \multicolumn{5}{c}{\textbf{Fixed Threshold F1 (0.5)}} \\
                \cmidrule(r){2-6} \cmidrule(r){7-11}
                
                 & Columbia & Coverage & CASIA1 & NIST16 & \small{COCOGlide} & Columbia & Coverage & CASIA1 & NIST16 & \small{COCOGlide} \\
                \midrule
                Mantra-Net \cite{Mantra-Net} & 0.650 & 0.486 & 0.320 & 0.225 & 0.673 & 0.508 & 0.317 & 0.180 & 0.172 & 0.516 \\
                SPAN \cite{SPAN} & 0.873 & 0.428 & 0.169 & 0.363 & 0.350 & 0.759 & 0.235 & 0.112 & 0.228 & 0.298 \\
                MVSS-Net \cite{MVSS-Net} & 0.781 & 0.659 & 0.650 & 0.372 & 0.642 & 0.729 & 0.514 & 0.528 & 0.320 & 0.486 \\
                PSCC-Net \cite{PSCC-Net} & 0.760 & 0.615 & 0.670 & 0.210 & 0.685 & 0.604 & 0.473 & 0.520 & 0.113 & 0.515 \\
                CAT-Net2 \cite{CAT-Net2} & 0.923 & 0.582 & 0.852 & 0.417 & 0.603 & 0.859 & 0.381 & 0.752 & 0.308 & 0.434 \\
                TruFor \cite{TruFor} & 0.914 & 0.735 & 0.822 & 0.470 & 0.720 & 0.859 & 0.600 & 0.737 & 0.399 & 0.523 \\
                UnionFor. \cite{UnionFormer} & 0.925 & 0.720 & 0.863 & 0.489 & 0.742 & 0.861 & 0.592 & 0.760 & 0.413 & 0.536 \\
                \textbf{ForgerySleuth} & \textbf{0.931} & \textbf{0.792} & \textbf{0.870} & \textbf{0.610} & \textbf{0.751} & \textbf{0.925} & \textbf{0.684} & \textbf{0.804} & \textbf{0.518} & \textbf{0.562}\\

                \bottomrule
            \end{tabular}
        }
    }
    \label{table:exp_localization_pretraining_F1}
    \vspace{-1em}
\end{table*}

\label{sec:experiment_setting}

% \noindent\textbf{Implementation Details.} We employ LLaVA-7B-v1-1 \cite{LLaVA} as the base multimodal LLM ($\mathcal{F}_{m}$) and use the ViT-H SAM \cite{SAM} backbone for the vision encoder ($\mathcal{F}_{v}$). For training, we utilize 2 NVIDIA 80GB A800 GPUs, with training scripts optimized by DeepSpeed \cite{DeepSpeed}, which helps reduce memory usage and accelerate training. We use the AdamW \cite{AdamW} optimizer, setting the learning rate to $0.0002$ with no weight decay. The learning rate is scheduled using WarmupDecayLR, with 100 warmup iterations. The weights for the text generation loss $\lambda_{txt}$ and mask loss $\lambda_{mask}$ are both set to $1.0$, while the BCE loss $\lambda_{bce}$ and DICE loss $\lambda_{dice}$ are weighted at $1.0$ and $0.2$, respectively. The batch size per device is $4$, with gradient accumulation steps set to $4$.

\noindent\textbf{Testing Dataset.} We utilize six publicly accessible test datasets, which are Columbia \cite{Columbia}, Coverage \cite{Coverage}, CASIA1 \cite{CASIA}, NIST16 \cite{NIST16}, IMD20 \cite{IMD20}, and COCOGlide \cite{TruFor}, to evaluate and compare our method with state-of-the-art methods thoroughly. To effectively evaluate the model's generalization capability, these datasets are ensured to be disjoint from the training data. Additionally, we use our ForgeryAnalysis-Eval dataset to assess the reasoning and analysis capabilities of the methods.

\vspace{0.05in}
\noindent\textbf{Evaluation Metrics.} Localizing the tampered regions at the pixel level is crucial in image manipulation detection. We follow established practices \cite{UnionFormer} by using optimal threshold and fixed threshold F1 scores and the threshold-independent Area Under the Curve (AUC) metric. To ensure fairness and precision in our comparative analysis, we refer to some results for other methods from \cite{UnionFormer, ObjectFormer}.

% Evaluating the novel reasoning analysis outputs presents a unique challenge, as it involves assessing the comprehension, reasoning, and correctness in generating text explanations. Inspired by previous work \cite{Vicuna, LLaVA}, we use GPT-4 as an automated evaluator to assess the reasoning performance of different models. Figure \ref{figure:sup_evaluation_gpt4} and Subsection \ref{sec:appendices_evaluation} in the \textit{Appendices} provide the prompt format and the evaluation criteria. Nevertheless, while GPT-4 has extensive evaluation capabilities, it is also susceptible to hallucinations. To address this limitation, we incorporate an additional metric, Semantic Textual Similarity (STS). We follow SBERT \cite{SBERT} and use STS to measure the similarity between the generated text and the ground-truth text in ForgeryAnalysis-Eval, where various models are utilized to calculate the similarity.

Evaluating the novel reasoning analysis outputs presents a unique challenge, as it involves assessing the comprehension, reasoning, and correctness in generating text explanations. To measure the similarity between the generated text and the ground-truth text, thereby reflecting its accuracy, we incorporate Semantic Textual Similarity (STS) metric. We follow SBERT \cite{SBERT} and use STS to measure the similarity in our ForgeryAnalysis-Eval, where various models are utilized to calculate the similarity. Additionally, inspired by previous work \cite{Vicuna, LLaVA}, we use GPT-4 as an automated evaluator to assess the reasoning performance of different models, which enables a more holistic evaluation of the generated analysis text, Figure \ref{figure:sup_evaluation_gpt4} and Section \ref{sec:appendices_evaluation} provide the prompt format and the evaluation criteria. We conduct evaluations twice and report the average performance.

% GPT-4 is responsible for reviewing the quality of the generated responses and assigning scores on a scale of 1 to 10 at various dimensions, where a higher score reflects better performance.

\subsection{Manipulation Detection Results}

% \begin{wraptable}{r}{0.51\textwidth}
\begin{table}[t]
    \centering
    \caption{Manipulation localization results of ForgerySleuth and SoTA methods, using pixel-level AUC as the evaluation metric.}
    \vspace{0.5em}
    \resizebox{0.65\linewidth}{!}{
        \setlength{\tabcolsep}{2mm}{
            \begin{tabular}{lccccc}
                \toprule
                \textbf{Method} & Columbia & Coverage & CASIA1 & NIST16 & IMD20 \\
                \midrule
                Mantra-Net \cite{Mantra-Net} & 0.824 & 0.819 & 0.817 & 0.795 & 0.748 \\
                SPAN \cite{SPAN} & 0.936 & 0.922 & 0.797 & 0.840 & 0.750 \\
                PSCC-Net \cite{PSCC-Net} & 0.982 & 0.847 & 0.829 & 0.855 & 0.806 \\
                ObjectFormer \cite{ObjectFormer} & 0.955 & 0.928 & 0.843 & 0.872 & 0.821 \\
                TruFor \cite{TruFor} & 0.947 & 0.925 & 0.957 & 0.877 & - \\
                UnionFormer \cite{UnionFormer} & 0.989 & 0.945 & \textbf{0.972} & 0.881 & 0.860 \\
                \textbf{ForgerySleuth} & \textbf{0.992} & \textbf{0.962} & 0.969 & \textbf{0.898} & \textbf{0.911} \\
                \bottomrule
            \end{tabular}
        }
    }
    \label{table:exp_localization_pretraining_AUC}
    \vspace{-0.5em}
\end{table}
% \end{wraptable}

The results in Table \ref{table:exp_localization_pretraining_F1} and Table \ref{table:exp_localization_pretraining_AUC}  demonstrate the performance of our ForgerySleuth and comparisons with SoTA methods for image manipulation localization, using pixel-level F1 scores and AUC metrics, respectively. Our method consistently achieves the highest or second-highest AUC. Regarding F1 scores, our approach surpasses other methods across all datasets, showcasing its reliability under both fixed and optimal thresholds. On challenging datasets like NIST16 and IMD20, ForgerySleuth outperforms UnionFormer by margins of 0.105 and 0.121 for fixed and optimal thresholds, respectively, which we believe is significant given its challenging nature. Furthermore, on the COCOGlide dataset, which features novel diffusion-based manipulations, our model also exceeds the UnionFormer. These significant improvements can be attributed to the capability to effectively capture both low-level trace features and high-level semantic inconsistencies, enabling it to detect even subtle generative manipulations. Overall, the results emphasize the generalization ability of ForgerySleuth.

\begin{table}[t]
        \centering
        \caption{Robust evaluation results of ForgerySleuth and existing methods using pixel-level AUC.}
        \vspace{0.5em}
        \resizebox{0.65\linewidth}{!}{
            \setlength{\tabcolsep}{2mm}{
                \begin{tabular}{lccccc}
                    \toprule
                    \textbf{Distortion} & SPAN & PSCC-Net & ObjectFor. & UnionFor. & \textbf{Ours} \\
                    \midrule
                    \textit{w/o} distortion & 0.8395 & 0.8547 & 0.8718 & 0.8813 & \textbf{0.8982} \\
                    Resize ($0.78\times$) & 0.8324 & 0.8529 & 0.8717 & 0.8726 & \textbf{0.8962}\\
                    Resize ($0.25\times$) & 0.8032 & 0.8501 & 0.8633 & 0.8719 & \textbf{0.8792}\\
                    GSBr ($k=3$) & 0.8310 & 0.8538 & 0.8597 & 0.8651 & \textbf{0.8863} \\
                    GSBr ($k=15$) & 0.7915 & 0.7993 & 0.8026 & 0.8430 & \textbf{0.8658} \\
                    GSN ($\sigma=3$) & 0.7517 & 0.7842 & 0.7958 & 0.8285 & \textbf{0.8452} \\
                    GSN ($\sigma=15$) & 0.6728 & 0.7665 & 0.7815 & 0.8057 & \textbf{0.8139} \\
                    JPEG ($q=100$) & 0.8359 & 0.8540 & 0.8637 & 0.8802 & \textbf{0.8974} \\
                    JPEG ($q=50$) & 0.8068 & 0.8537 & 0.8624 & 0.8797 & \textbf{0.8839} \\
                    \bottomrule
                \end{tabular}
            }
        }
        \label{table:exp_localization_robust}
    \vspace{-0.5em}

\end{table}

\vspace{0.05in}
\noindent\textbf{Robustness Evaluation.} We apply several image distortions to the NIST16 dataset following \cite{UnionFormer, ObjectFormer} to evaluate the robustness of our method and compare the results with other methods. The distortions included 1) resizing images to different scales, 2) applying Gaussian blur with different kernel sizes $k$, 3) adding Gaussian noise with various standard deviation $\sigma$, and 4) compressing images using JPEG with different quality factors $q$. The results in Table \ref{table:exp_localization_robust} show that our model consistently outperforms other methods across all types of distortions. This improvement in robustness stems from the ability to identify high-level semantic anomalies rather than relying solely on low-level statistical features that are more susceptible to distortions.

\subsection{Forgery Analysis Results}
\label{subsec:forgery_analysis}

We compare our forgery analysis results with several M-LLMs, including GPT-4o, Qwen2-VL \cite{Qwen2-VL}, and LISA \cite{LISA}. Additionally, we perform LoRA fine-tuning on the LISA using the ForgeryAnalysis-SFT to provide a more comprehensive comparison. We incorporate STS to measure the similarity between the generated text and the ground-truth text in ForgeryAnalysis-Eval. The results, shown in Table \ref{table:exp_analysis_sts},  confirm that our ForgerySleuth consistently outperforms other methods.

We also leverage GPT-4 as an evaluator based on the ForgeryAnalysis-Eval dataset to assess the quality of text analysis and reasoning. We collect answers from each M-LLM, and GPT-4 assigns a score from 1 to 10 for each response. Beyond scoring, GPT-4 provides detailed explanations for its ratings. We also report the recall rate to directly reflect the ability to identify tampered images.

\begin{figure}[t]
    \centering

    \begin{minipage}[t]{0.675\textwidth}
          \centering
          \includegraphics[width=\linewidth]{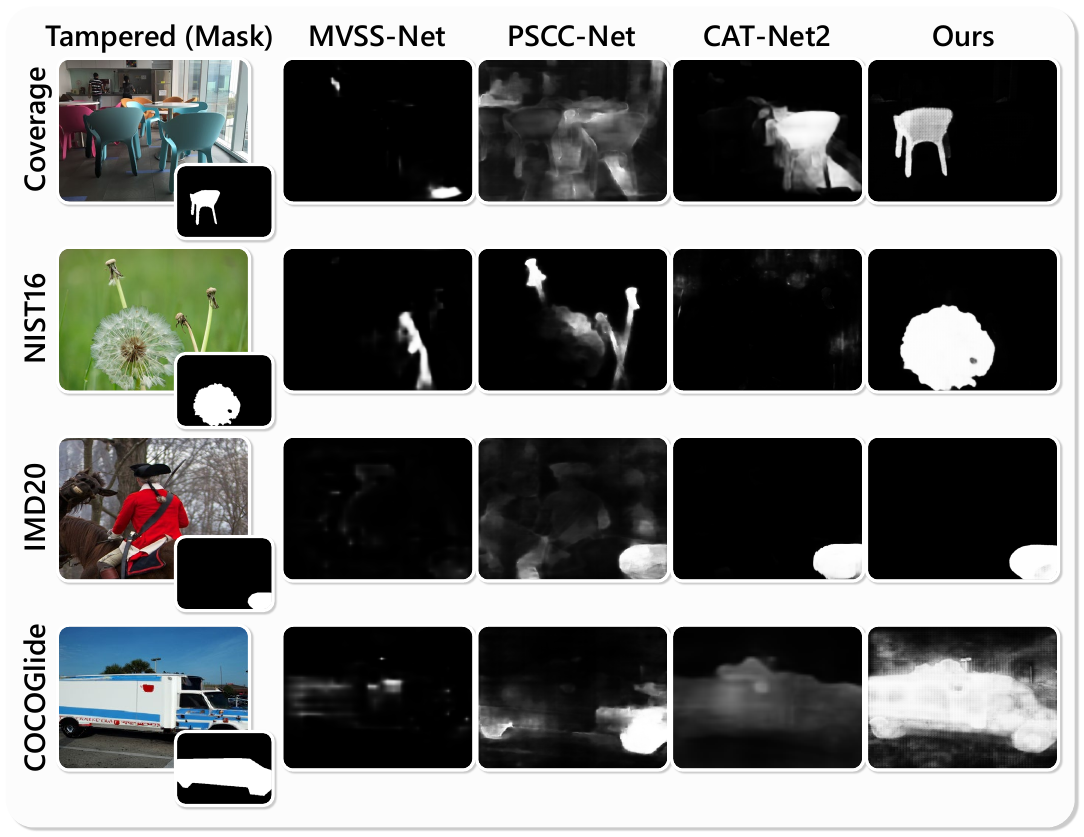}
          \caption{Visualization results comparing ForgerySleuth with existing methods. The examples are taken from various datasets.}
          \label{figure:visualization}
    \end{minipage}\hfill
    \begin{minipage}[t]{0.314\textwidth}
        \centering
        \includegraphics[width=\linewidth]{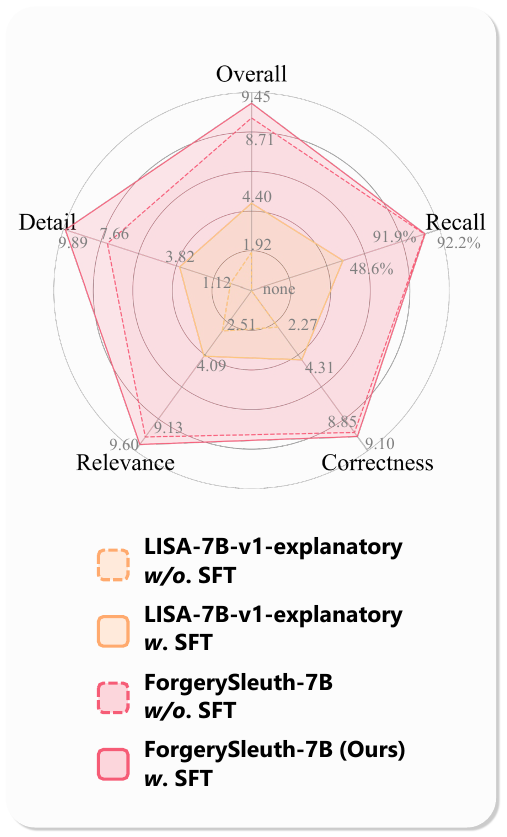}
        \caption{Forgery analysis results of the ablation study.}
        \label{figure:radar}
    \end{minipage}

\vspace{-1em}
    
\end{figure}

Figure \ref{figure:teaser} (b) presents the scores of different models without additional fine-tuning across various evaluation dimensions. The existing models exhibit low recall rates, and the overall evaluation suggests that they struggle to identify manipulations and provide accurate analyses. Figure \ref{figure:radar} shows the scores of LISA and ForgerySleuth after SFT, along with the versions without SFT. ForgerySleuth shows an improvement of 5.05 in the overall score compared to LISA, further demonstrating the effectiveness of our design targeted specifically for the IMD task. Furthermore, using the SFT dataset results in performance gains for both methods, indicating the quality of the ForgeryAnalysis dataset.

\subsection{Ablation Study}

We conduct an extensive ablation study to analyze the effect of our ForgeryAnalysis dataset and each component and setting within ForgerySleuth framework. We report the pixel-level localization performance on IMD datasets using the F1 score. We use ForgeryAnalysis-Eval to evaluate the quality of the reasoning text, following the scoring criteria described in Section \ref{subsec:forgery_analysis}.

\vspace{0.05in}
\noindent\textbf{Contribution of ForgeryAnalysis.} The experimental results in Table \ref{table:exp_data_ablation} show the performance drop when ForgeryAnalysis-PT data is excluded during the pre-training phase, demonstrating the importance of large-scale data. Using ForgeryAnalysis-PT, the model achieves strong performance on the IMD task, even without the final stage of supervised fine-tuning with ForgeryAnalysis-SFT data. However, this final fine-tuning further enhances the quality of the tampering analysis text.

\begin{table}[t]
    \centering

    \begin{minipage}[c]{0.5\textwidth}
        \centering
        \caption{Forgery analysis results of ForgerySleuth and SoTA methods, evaluated using STS.}
        \vspace{0.5em}
        \resizebox{\linewidth}{!}{
            \setlength{\tabcolsep}{1mm}{
                \begin{tabular}{lc|ccccc}
                    \toprule
                    \textbf{Model} & \textbf{Dim} & \textbf{Ours} & GPT-4o & QWen2 & QWen & LISA \\
                    \midrule
                    \small{MiniLM-L6-v2} & 384 & \textbf{0.926} & 0.725 & 0.595 & 0.551 & 0.313 \\
                    \small{MiniLM-L12-v2} & 384 & \textbf{0.919} & 0.645 & 0.505 & 0.475 & 0.350 \\
                    \small{mpnet-base-v2} & 768 & \textbf{0.961} & 0.724 & 0.635 & 0.546 & 0.401 \\
                    \bottomrule
                \end{tabular}
            }
        }
        \label{table:exp_analysis_sts}
    \end{minipage}\hfill
    \begin{minipage}[c]{0.48\textwidth}
        \centering
        \caption{Ablation study on different parts of ForgeryAnalysis.}
        \vspace{0.5em}
        \resizebox{\linewidth}{!}{
            \setlength{\tabcolsep}{0.5mm}{
                \begin{tabular}{cccccc}
                    \toprule
                    \textbf{Data} & NIST16 & IMD20 & \footnotesize{COCOGlide} & ForgeryA. \\
                    \midrule
                    ForgeryA.-SFT & 0.191 & 0.094 & 0.152 & 5.09\\
                    ForgeryA.-PT & 0.516 & 0.705 & \textbf{0.571} & 8.71\\
                    ForgeryA.-PT\&SFT & \textbf{0.518} & \textbf{0.710} & 0.562 & \textbf{9.45}\\
                    \bottomrule
                \end{tabular}
            }
        }
        \label{table:exp_data_ablation}
    \end{minipage}

\vspace{-1em}

\end{table}

% \begin{wraptable}[10]{r}{0.5\textwidth}
\begin{table}[t]
    \centering
    \vspace{0.5em}
    \caption{Ablation study on different modules and settings, using pixel-level F1 with fixed threshold 0.5 as the evaluation metric.}
    \vspace{0.5em}
    \resizebox{0.65\linewidth}{!}{
        \setlength{\tabcolsep}{2.8mm}{
            \begin{tabular}{lcccc}
                \toprule
                \textbf{Setting} & CASIA1 & NIST16 & IMD20 & COCOGlide \\
                \midrule
                1. \textit{w/o.} Trace Enc. $\mathcal{F}_{t}$ & 0.637 & 0.323 & 0.622 & 0.395 \\
                2. \textit{w/o.} Fusion Mechanism & 0.628 & 0.463 & 0.649 & 0.513\\
                \midrule
                3. \textit{w.} $\mathcal{F}_{v}$ Trainable & 0.755 & 0.451 & 0.716 & 0.526\\
                4. \textit{w.} $\mathcal{F}_{v}$ LoRA-ft. & 0.766 & 0.493 & \textbf{0.731} & 0.547 \\
                \midrule
                \textbf{ForgerySleuth} & \textbf{0.804} & \textbf{0.518} & 0.710 & \textbf{0.562} \\
                \bottomrule
            \end{tabular}
        }
    }
    \label{table:exp_localization_ablation}
\end{table}
% \end{wraptable}

\vspace{0.05in}
\noindent\textbf{Effect of Designed Components.} As shown in Table \ref{table:exp_localization_ablation}, removing the trace encoder $\mathcal{F}_{t}$ significantly degrades performance across all datasets, confirming its critical contribution in capturing low-level trace features and mitigating hallucinations in the M-LLM. At the same time, the performance drop observed when the fusion mechanism is excluded highlights the effectiveness of our fusion strategy in integrating multimodal and trace-based information.

\vspace{0.05in}
\noindent\textbf{Impact of Training Strategies.} In our comparison of various training strategies for the visual encoder $\mathcal{F}_{v}$, including trainable and LoRA fine-tuned, we discover that the best performance is achieved by keeping the encoder completely frozen. This could be attributed to the fact that trainable and LoRA fine-tuned strategies slightly diminish the generalization ability of the original SAM. 

% Furthermore, these strategies may result in the loss of the inherent capability to represent the normal features of authentic images.

\subsection{Qualitative Results}

Figure \ref{figure:visualization} showcases pixel-level localization results across different datasets, comparing our framework with other SoTA methods. The masks are displayed without binarization to provide a more detailed view of the localization capability. Our method consistently delivers more precise tampered region detection with higher confidence across various types of manipulation. More detailed examples of clues and analysis provided by ForgerySleuth can be found in Section \ref{sec:appendices_cases}. Across different types of manipulation, relevant clues and high-quality analysis demonstrate the effectiveness of M-LLM in capturing high-level semantic anomalies.

\section{Conclusion}
\label{sec:conclusion}

In this work, we explored the potential of multimodal large language models in the image manipulation detection task. The proposed ForgerySleuth integrates M-LLMs with a trace encoder, allowing the model to utilize world knowledge to detect high-level semantic anomalies while effectively capturing low-level forgery traces. Additionally, we introduced a vision decoder with a fusion mechanism to integrate different features, ultimately producing precise segmentation masks. We also proposed a supervised fine-tuning dataset, ForgeryAnalysis, specifically designed for the IMD task. Each entry was initially generated by GPT-4o using a novel Chain-of-Clues prompt and then reviewed and refined by experts. Furthermore, we developed a data engine based on this dataset to automate forgery analysis, facilitating the creation of a larger-scale ForgeryAnalysis-PT dataset for pre-training purposes. A discussion of limitations can be found in Section \ref{sec:limitation}. We have already made the resources publicly available, including the data, code, and weights, to provide resources for advancing the field.

\vspace{0.05in}
\noindent\textbf{Acknowledgment.} This work was supported in part by the National Natural Science Foundation of China (Grant 62472098). We would like to express our gratitude to Ruirui Tu and Xu Han for their valuable assistance with data annotation, which significantly contributed to the development of the ForgeryAnalysis dataset.

\bibliographystyle{IEEEtran}
\bibliography{main}

\clearpage
\appendix

\section{ForgeryAnalysis Dataset}

\label{sec:appendices_dataset}

\subsection{ForgeryAnalysis Data}

We utilize the advanced M-LLM, GPT-4o, to generate the initial clue analyses, carefully designing prompts to ensure GPT-4o provides accurate and detailed responses. First, we inform GPT-4o of its role as an assistant, clearly outlining the levels of clues, along with specific examples, and specifying the task it needs to complete. The detailed prompt is as follows:

\vspace{0.05in}
\noindent {\usefont{T1}{DejaVuSans-TLF}{m}{} [ROLE] You are a rigorous and responsible image tampering (altering) detection expert. You can detect whether an image has been tampered with, localize the exact tampered region, and analyze your detection decision according to tampering clues of different levels. These clues include but are not limited to low-level pixel statistics (such as noise, color, textural, sharpness, and AI-generation fingerprint), middle-level visual defects (such as traces of tampered region or boundary, lighting inconsistency, perspective relationships, and physical constraints), and high-level semantic anomalies (such as content contrary to common sense, inciting and misleading content), etc. Altering operations could be divided into types, including ``splice'', ``copy-move'', ``remove'', and ``AI-generate'', leaving different clues that you should consider.}

\noindent {\usefont{T1}{DejaVuSans-TLF}{m}{} [TASK] Now, your task is to provide analysis. Please note that in real detection scenarios, you cannot know in advance whether an image has been tampered with and the specific tampered region. However, now I will tell you this information to help you conduct a more rigorous and accurate analysis based on this. There is no need to include all aspects and views in your analysis, give some of your most confident points.}
\vspace{0.05in}

\begin{figure*}[tbh]
  \centering
  \includegraphics[width=\linewidth]{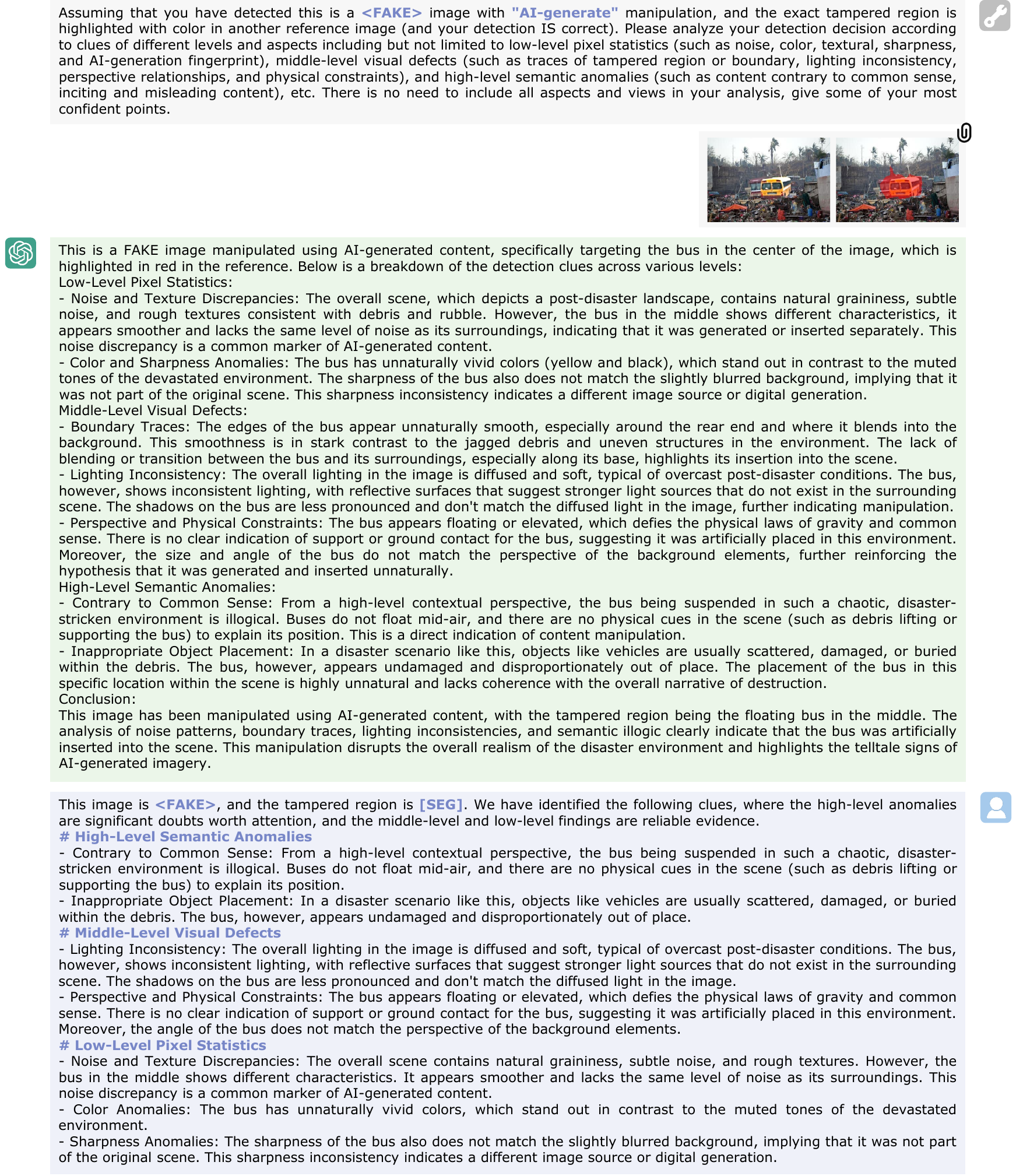}
  \caption{Illustration of the ForgeryAnalysis dataset creation process. The data is initially generated by GPT-4o, then revised by experts to ensure the accuracy of the analysis, with clues organized in the Chain-of-Clues (CoC) format.}
  \label{figure:sup_engine_gpt4o}
\end{figure*}

The details of each conversation round are illustrated in Figure \ref{figure:sup_engine_gpt4o}. In each round, we provide two images: the tampered image to be analyzed and a reference image with the tampered region highlighted. The prompt includes the \texttt{<FAKE>} token to indicate that the image is manipulated, specific tampering types to help the model focus on relevant clues, and a structured clue analysis format. Based on the response of GPT-4o, experts then conduct further revisions. The experts carefully review vague and incorrect statements that may arise from hallucinations, removing irrelevant clues and evidence. They also reorganize the clues into the Chain-of-Clues structure, which guides the reasoning process. This begins with ``unveiling doubts'' using high-level clues and continues with ``pinpointing evidence'' using middle-level and low-level features. The experts also check special tokens, such as \texttt{<FAKE>} and \texttt{[SEG]}, to meet the requirements for subsequent model training.

\subsection{ForgeryAnalyst Engine}

\label{sec:appendices_engine}

It is worth noting that the M-LLMs used in the data engine ForgeryAnalyst and the detection framework ForgerySleuth are different and independent. Although both M-LLMs are designed to analyze forgery and produce text-based clue analyses, their inputs and tasks are distinct. ForgeryAnalyst receives input that includes explicit information about the tampered region (highlighted to indicate tampering) with the goal of generating more precise and comprehensive clue analyses to construct pre-training data. In contrast, ForgerySleuth takes an image to be analyzed, aiming for the M-LLM to identify high-level semantic anomalies for detecting tampered regions. Furthermore, we design different prompts tailored to these two specific tasks. In our experiments, ForgeryAnalyst employs LLaVA-v1.5-13B, while the MLLM in ForgerySleuth uses LLaVA-7B-v1-1, balancing performance and efficiency.

We also designed a dedicated prompt for the data engine. In addition to indicating the image type as \texttt{<FAKE>} and specifying the tampering type [MANIPULATION-TYPE], we provide detailed instructions and examples of the Chain-of-Clues (CoC), as well as the required output data format. The specific prompt format is as follows:

\vspace{0.05in}
\noindent {\usefont{T1}{DejaVuSans-TLF}{m}{} You are a rigorous and responsible image tampering (altering) detection expert. You can localize the exact tampered region and analyze your detection decision according to tampering clues at different levels. Assuming that you have detected this is a <FAKE> image and the manipulation type is [MANIPULATION-TYPE], the exact tampered region boundary is highlighted with color in this image (and your detection IS correct).}

\noindent {\usefont{T1}{DejaVuSans-TLF}{m}{} Please provide the chain-of-clues supporting your detection decision in the following style: \# high-level semantic anomalies (such as content contrary to common sense, inciting and misleading content), \# middle-level visual defects (such as traces of tampered region or boundary, lighting inconsistency, perspective relationships, and physical constraints) and \# low-level pixel statistics (such as noise, color, textural, sharpness, and AI-generation fingerprint), where the high-level anomalies are significant doubts worth attention, and the middle-level and low-level findings are reliable evidence.}
\vspace{0.05in}

\subsection{Statistics}

\begin{table}[t]
    \centering
    \caption{Statistic of our ForgeryAnalysis Dataset.}
    \vspace{0.5em}
    \resizebox{0.6\linewidth}{!}{
        \setlength{\tabcolsep}{3mm}{
            \begin{tabular}{|l|c|c|c|}
                \hline
                \textbf{Dataset} & \textbf{Split} & \textbf{Count} & \textbf{Sum} \\
                
                \hline
                
                \multirow{2}[1]{*}{ForgeryAnalysis} & ForgeryAnalysis-Eval & 618 & \multirow{2}[1]{*}{2,370} \\

                \cline{2-3}
                 & ForgeryAnalysis-SFT & 1752 & \\
                
                \hline                
                
                ForgeryAnalysis-PT & - & 50k & 50k\\
                
                \hline
            \end{tabular}
        }
    }
    \label{table:sta_dataset}
\end{table}

\begin{table}[t]
    \centering
    \caption{Overview of public dataset utilized in the construction of ForgeryAnalysis and for evaluation.}
    \vspace{0.5em}
    \resizebox{\linewidth}{!}{
        \setlength{\tabcolsep}{2mm}{
            \begin{tabular}{|l|c|c|c|c|c|}
                \hline
                \textbf{Dataset} & \textbf{Usage} & \textbf{Authentic} & \textbf{Tampered} & \textbf{Manipulation Types} & \textbf{Source} \\
                \hline
                MIML [34] & Train, Construct & 11,142 & 123,150 & Manual Editing & PSBattles \\
                CASIA2 [35] & Train, Construct & 7,491 & 5,063 & Splice, Copy-move & Corel \\
                DEFACTO [36] & Train, Construct & - & 149,587 & Splice, Copy-move, AIGC (GAN) & MS COCO \\
                AutoSplice [37] & Train, Construct & 2,273 & 3,621 & AIGC & Visual News \\
                \hline
                Columbia [55] & Eval & 183 & 180 & Splice & - \\
                Coverage [56] & Eval & 100 & 100 & Copy-move & - \\
                CASIA1 [35] & Eval & 800 & 920 & Splice, Copy-move & Corel \\
                NIST16 [57] & Eval & - & 564 & Splice, Copy-move & - \\
                IMD20 [58] & Eval & 414 & 2,010 & Manual Editing & - \\
                COCOGlide [30] & Eval & 512 & 512 & AIGC (Diffusion) & - \\
                \hline
            \end{tabular}
        }
    }
    \label{table:sta_public_dataset}
\end{table}

Table \ref{table:sta_dataset} presents the data statistics of the ForgeryAnalysis dataset. ForgeryAnalysis-Eval and ForgeryAnalysis-SFT are initially generated by GPT-4o and fully revised by experts. They are used for evaluating the quality of manipulation analysis generated by the M-LLMs and for the final supervised fine-tuning, respectively. ForgeryAnalysis-PT is automatically constructed by our proposed data engine, ForgeryAnalyst, maintaining consistency in data format with the other subsets. Table \ref{table:sta_public_dataset} provides detailed statistics on the scale, source, and manipulation types from each public dataset utilized in the construction of ForgeryAnalysis and for evaluation.

\subsection{Cases}

\label{sec:appendices_cases}

\begin{figure*}[t]
  \centering
  \includegraphics[width=\linewidth]{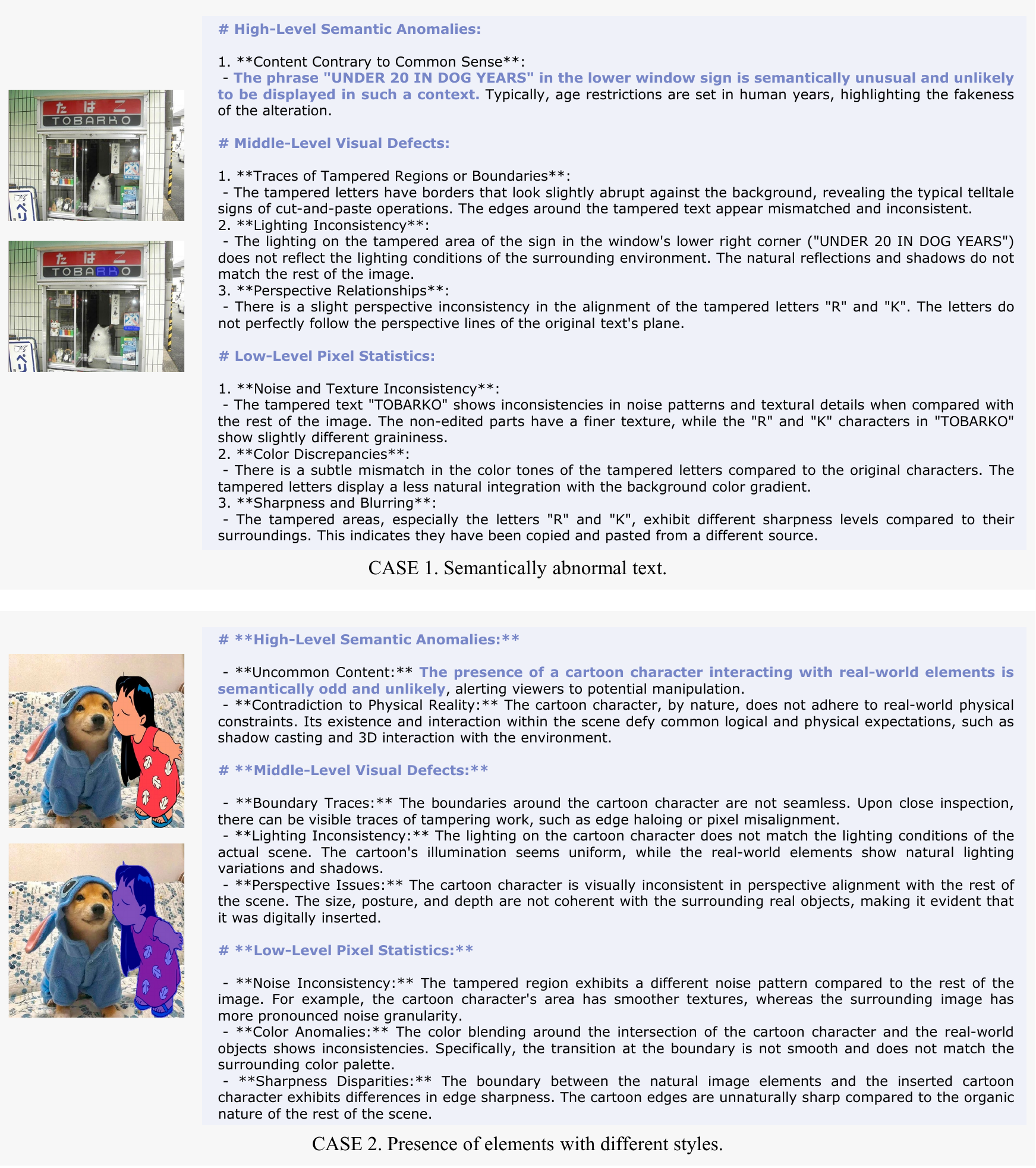}
  \caption{Examples from the ForgeryAnalysis-Eval dataset. The data is initially generated by GPT-4o and then cross-revised by multiple experts. The manipulation type for these images is ``splice''.}
  \label{figure:sup_fa_eval_splice}
\end{figure*}

\begin{figure*}[t]
  \centering
  \includegraphics[width=\linewidth]{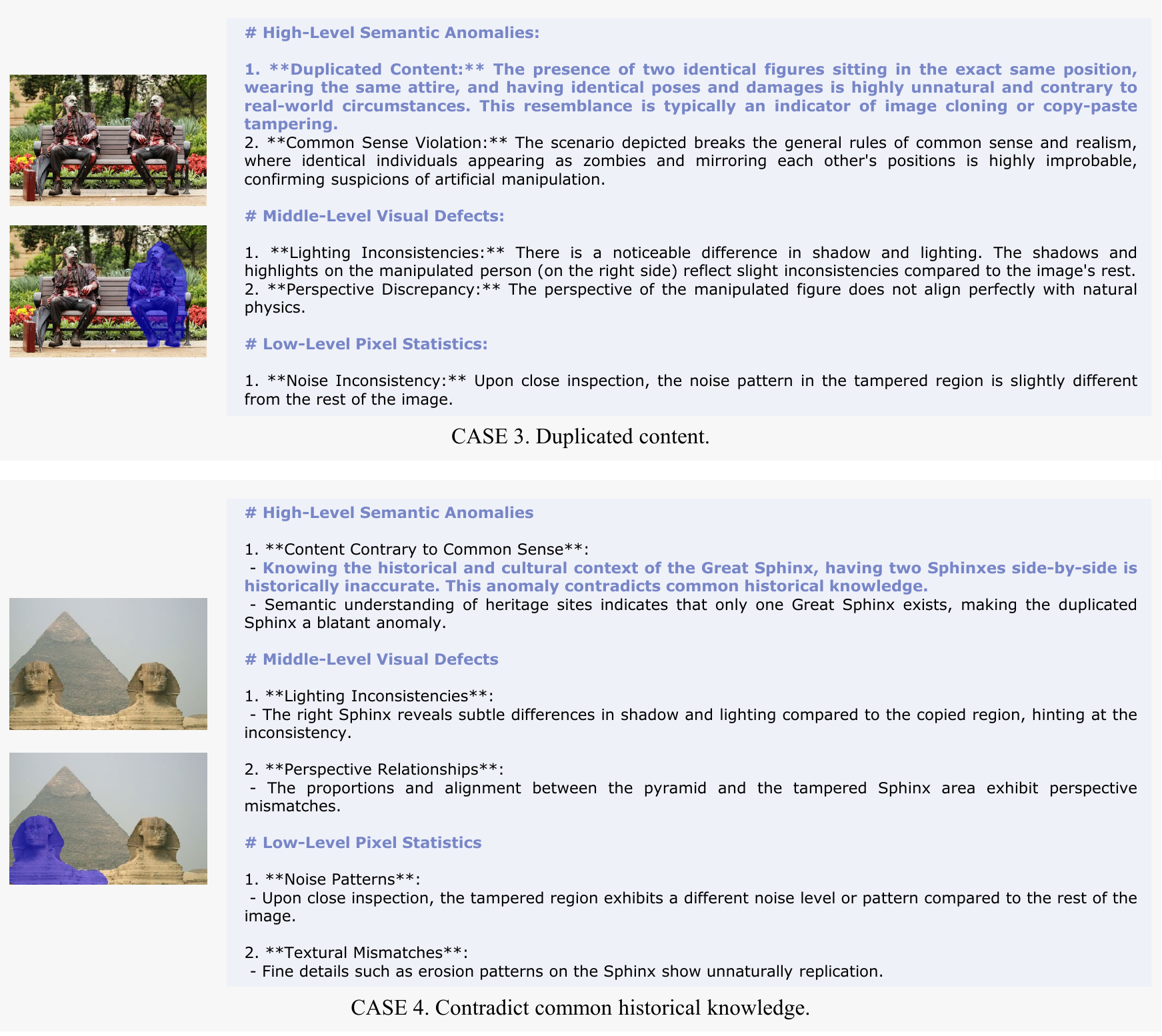}
  \caption{Examples from the ForgeryAnalysis-Eval dataset. The data is initially generated by GPT-4o and then cross-revised by multiple experts. The manipulation type for these images is ``copy-move''.}
  \label{figure:sup_fa_eval_cpmv}
\end{figure*}

\begin{figure*}[t]
  \centering
  \includegraphics[width=\linewidth]{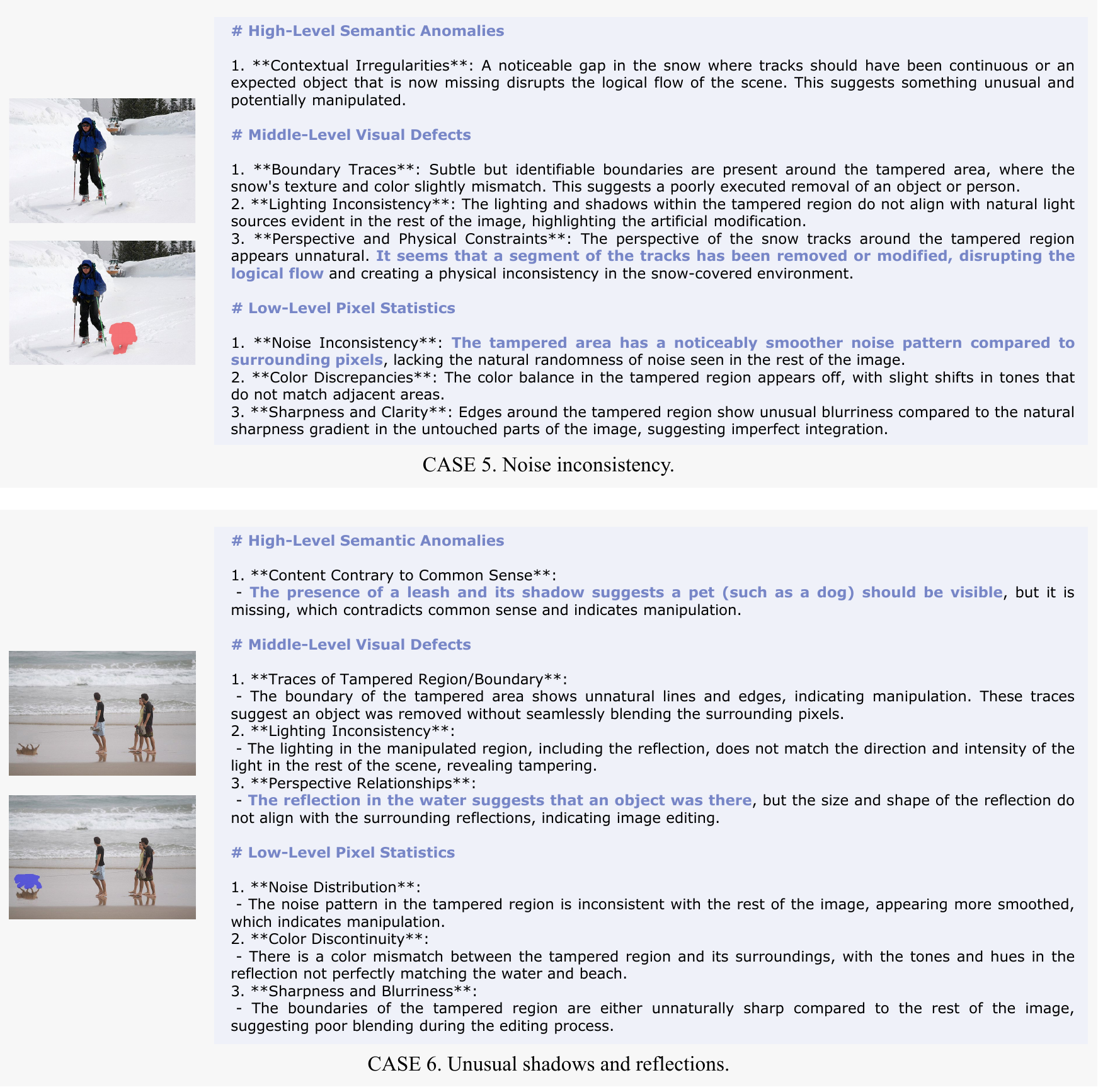}
  \caption{Examples from the ForgeryAnalysis-Eval dataset. The data is initially generated by GPT-4o and then cross-revised by multiple experts. The manipulation type for these images is ``remove''.}
  \label{figure:sup_fa_eval_remove}
\end{figure*}

\begin{figure*}[t]
  \centering
  \includegraphics[width=\linewidth]{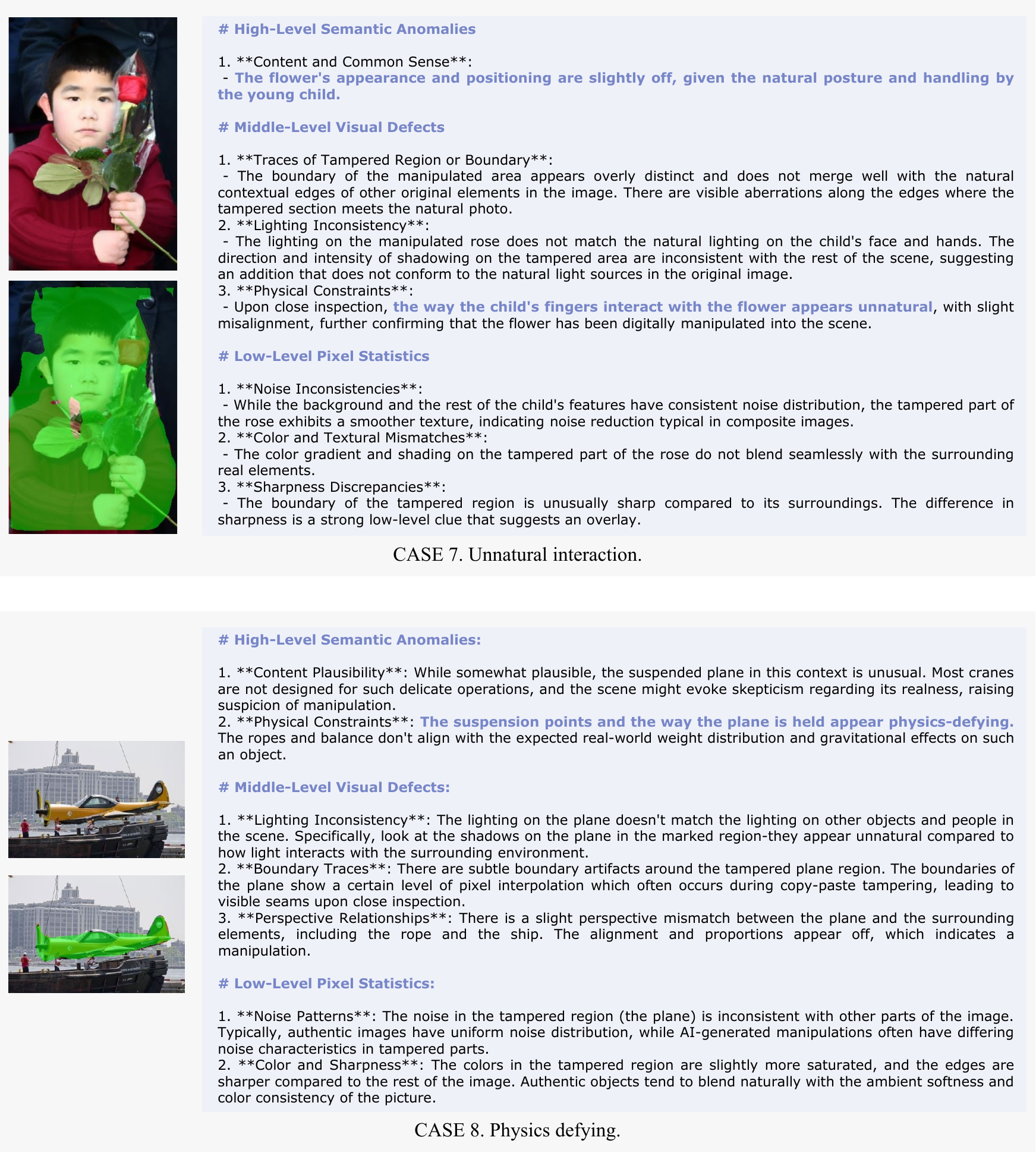}
  \caption{Examples from the ForgeryAnalysis-Eval dataset. The data is initially generated by GPT-4o and then cross-revised by multiple experts. The manipulation type for these images is ``AI-generate''.}
  \label{figure:sup_fa_eval_aigen}
\end{figure*}

\begin{figure*}[t]
  \centering
  \includegraphics[width=\linewidth]{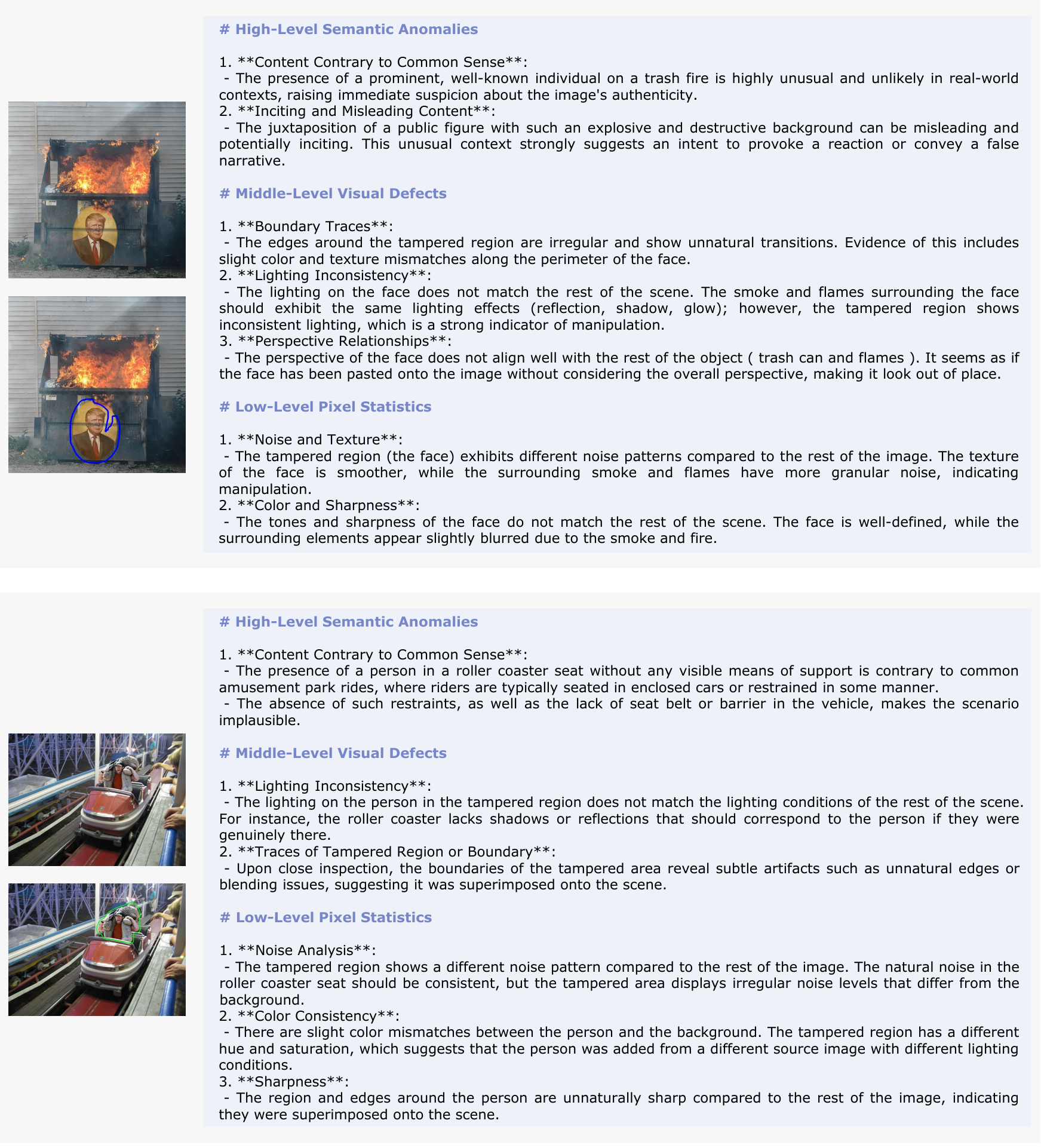}
  \caption{Examples from the ForgeryAnalysis-PT dataset. The data is automatically generated by our data engine ForgeryAnalyst.}
  \label{figure:sup_fa_pt}
\end{figure*}

Figures \ref{figure:sup_fa_eval_splice}, \ref{figure:sup_fa_eval_cpmv}, \ref{figure:sup_fa_eval_remove} and \ref{figure:sup_fa_eval_aigen} present examples of analysis texts from ForgeryAnalysis-Eval for four different tampering types. These diverse cases highlight the variations in detectable clues across different tampering types and illustrate how varying levels of clues support manipulation detection. The analyses in ForgeryAnalysis-Eval are cross-checked by multiple experts to ensure comprehensive and accurate clues. Figure \ref{figure:sup_fa_pt} shows analysis texts in ForgeryAnalysis-PT. Although the subset is automatically generated by the data engine, it also provides precise descriptions and analyses of the tampered regions.

\section{ForgerySleuth}

\subsection{Framework Details}

\label{sec:details}

\vspace{0.05in}
\noindent\textbf{Trace Encoder.}
Considering that the vision backbone (ViT-H SAM backbone) is pre-trained on tasks highly correlated with semantics and remains frozen during the fine-tuning of our framework, the semantic-agnostic features widely used in IMD tasks \cite{Mantra-Net, SPAN, MVSS-Net, SAFL-Net} cannot be effectively leveraged. We propose an independent trace encoder $\mathcal{F}_{t}$, equipped with a noise enhancement module to focus on low-level features and provide more reliable tampering evidence. Specifically, the noise enhancement module, positioned at the front of the encoder, uses constrained convolutions \cite{Bayar} to compute local differences, extract noise features, and suppress image content. The convolution kernel constraints are defined as follows:
\begin{equation}
    \left\{\begin{aligned}\omega_{(0,0)}&=1, \\ \sum_{(m,n)}\omega_{(m,n)}&=0, \end{aligned} \right.
\end{equation}
where $(m, n)$ denotes the spatial index of the values within the convolution kernel, with $(0, 0)$ positioned at the center. The constrained convolutions are still trainable, allowing them to learn manipulation features more adaptively than fully fixed-parameter noise extractors. The extracted noise features further enhance the original features by residual connections. The encoder employs a ViT-B architecture and all parameters are fine-tuned during training.

In summary, we employ the trace encoder $\mathcal{F}_{t}$ to capture low-level manipulation features, leveraging constrained convolutions within the noise enhancement module to achieve this. Meanwhile, the vision backbone $\mathcal{F}_{v}$ utilizes pre-trained parameters from SAM, which is widely recognized for its ability to capture dense visual content features, thereby supporting more precise segmentation.

\vspace{0.05in}
\noindent\textbf{Fusion Mechanism.} 
To integrate the image content embedding $\mathbf{f}_{c}$, trace embedding $\mathbf{f}_{t}$, and the LLM output tokens $\mathbf{h}_\text{seg}$ obtained from the M-LLM, we take inspiration from Transformer segmentation models \cite{DETR, MaskFormer, SAM} and design a vision decoder with a fusion attention mechanism, illustrated in Figure \ref{figure:framework} (b). The mechanism consists of three layers, with each layer performing four steps: self-attention on the LLM output tokens or upgraded tokens, cross-attention from tokens (as queries) to the trace or image embeddings, point-wise MLP, and cross-attention from the trace or image embeddings (as queries) back to the tokens. The first layer of the module computes attention between anomalies in the LLM output tokens and traces in the trace embeddings, facilitating the organization and pinpointing of clues. The subsequent layers focus on attention between refined clues in the upgraded tokens and content in the image embeddings, enabling more precise segmentation of the tampered regions.

% We apologize for the imprecise terms about the fusion mechanism in the original manuscripts, which may have caused some confusion. We have corrected these terms, such as updating ``prompt tokens'' to ``LLM output tokens'', and will ensure consistency between the main text and supplementary material. Along with Figure \ref{figure:fusion} and the supplementary details, the updated version provides more precise descriptions of the fusion mechanism.

\begin{table*}[t]
    \centering
    \caption{Manipulation localization results comparing ForgerySleuth with SoTA methods.}
    % for optimal and fixed thresholds (0.5)
    \resizebox{\linewidth}{!}{
        \setlength{\tabcolsep}{1mm}{
            \begin{tabular}{lcccccccccc}
                \toprule
                \multirow{2}[1]{*}{\textbf{Method}} & \multicolumn{5}{c}{\textbf{Optimal Threshold F1}} & \multicolumn{5}{c}{\textbf{Fixed Threshold F1 (0.5)}} \\
                \cmidrule(r){2-6} \cmidrule(r){7-11}
                
                 & Columbia & Coverage & CASIA1 & NIST16 & \small{COCOGlide} & Columbia & Coverage & CASIA1 & NIST16 & \small{COCOGlide} \\
                \midrule
                Mantra-Net \cite{Mantra-Net} & 0.650 & 0.486 & 0.320 & 0.225 & 0.673 & 0.508 & 0.317 & 0.180 & 0.172 & 0.516 \\
                SPAN \cite{SPAN} & 0.873 & 0.428 & 0.169 & 0.363 & 0.350 & 0.759 & 0.235 & 0.112 & 0.228 & 0.298 \\
                MVSS-Net \cite{MVSS-Net} & 0.781 & 0.659 & 0.650 & 0.372 & 0.642 & 0.729 & 0.514 & 0.528 & 0.320 & 0.486 \\
                PSCC-Net \cite{PSCC-Net} & 0.760 & 0.615 & 0.670 & 0.210 & 0.685 & 0.604 & 0.473 & 0.520 & 0.113 & 0.515 \\
                CAT-Net2 \cite{CAT-Net2} & 0.923 & 0.582 & 0.852 & 0.417 & 0.603 & 0.859 & 0.381 & 0.752 & 0.308 & 0.434 \\
                TruFor \cite{TruFor} & 0.914 & 0.735 & 0.822 & 0.470 & 0.720 & 0.859 & 0.600 & 0.737 & 0.399 & 0.523 \\
                UnionFor. \cite{UnionFormer} & 0.925 & 0.720 & 0.863 & 0.489 & 0.742 & 0.861 & 0.592 & 0.760 & 0.413 & 0.536 \\
                \rowcolor{gray!15}
                FakeShield \cite{FakeShield} & 0.306 & 0.085 & 0.620 & 0.119 & 0.659 & 0.285 & 0.052 & 0.566 & 0.099 & 0.536 \\
                \rowcolor{gray!15}
                - only \textit{TP samples} & 0.874 & 0.470 & 0.696 & 0.514 & 0.659 & 0.813 & 0.289 & 0.635 & 0.431 & 0.536 \\
                \rowcolor{gray!30}
                \textbf{ForgerySleuth} & \textbf{0.931} & \textbf{0.792} & \textbf{0.870} & \textbf{0.610} & \textbf{0.751} & \textbf{0.925} & \textbf{0.684} & \textbf{0.804} & \textbf{0.518} & \textbf{0.562}\\

                \bottomrule
            \end{tabular}
        }
    }
    \label{table_supp:exp_localization_pretraining_F1}
\end{table*}

\begin{table}[t]
    \centering
    \caption{Manipulation localization results of ForgerySleuth and SoTA methods, using pixel-level AUC as the evaluation metric.}
    \vspace{0.5em}
    \resizebox{0.65\linewidth}{!}{
        \setlength{\tabcolsep}{2mm}{
            \begin{tabular}{lccccc}
                \toprule
                \textbf{Method} & Columbia & Coverage & CASIA1 & NIST16 & IMD20 \\
                \midrule
                Mantra-Net \cite{Mantra-Net} & 0.824 & 0.819 & 0.817 & 0.795 & 0.748 \\
                SPAN \cite{SPAN} & 0.936 & 0.922 & 0.797 & 0.840 & 0.750 \\
                PSCC-Net \cite{PSCC-Net} & 0.982 & 0.847 & 0.829 & 0.855 & 0.806 \\
                ObjectFor. \cite{ObjectFormer} & 0.955 & 0.928 & 0.843 & 0.872 & 0.821 \\
                TruFor \cite{TruFor} & 0.947 & 0.925 & 0.957 & 0.877 & - \\
                UnionFor. \cite{UnionFormer} & 0.989 & 0.945 & \textbf{0.972} & 0.881 & 0.860 \\
                
                \rowcolor{gray!15}
                FakeShield \cite{FakeShield} & 0.323 & 0.137 & 0.787 & 0.185 & 0.780 \\
                \rowcolor{gray!15}
                - only \textit{TP samples} & 0.924 & 0.761 & 0.883 & 0.801 & 0.868 \\
                \rowcolor{gray!30}
                \textbf{ForgerySleuth} & \textbf{0.992} & \textbf{0.962} & 0.969 & \textbf{0.898} & \textbf{0.911} \\
                \bottomrule
            \end{tabular}
        }
    }
    \label{table_supp:exp_localization_pretraining_AUC}
\end{table}

\section{Comparison with FakeShield}

\label{sec:fakeshield}

\subsection{Details}

FakeShield \cite{FakeShield} is a concurrent work with the same motivation as ours. It proposes a multimodal large model for image manipulation detection. This further reinforces the significance of providing a reasonable forgery analysis, alleviating the explainability issues in existing image manipulation detection methods. However, there are some fundamental differences between FakeShield and our approach.

From a data perspective, FakeShield relies on GPT-4o to generate tampering analysis texts, constructing ``image-mask-description'' triplets for training and evaluation. However, this process presents certain challenges: First, the generated texts lack structured reasoning, as FakeShield does not impose explicit guidance on how the detected clues is organized and analysis data is formulated. Second, reliance on GPT-4o alone makes data prone to hallucinations, potentially introducing unreliable or misleading explanations.

In contrast, we propose a three-level Chain-of-Clues (CoC) structure, which enables a more structured and interpretable analysis. Our design is based on two key insights: 1) Previous research \cite{wang2009survey, zheng2019survey} suggests that forensic clues are naturally organized hierarchically. 2) Studies on Chain-of-Thought (CoT) prompting \cite{CoT, ToT} have demonstrated its effectiveness in improving reasoning within large language models. Inspired by these findings, we structure the tampering analysis data into a progressive reasoning process, encouraging the model to identify, refine, and interpret forensic clues at different levels. Additionally, to reduce hallucinations and improve reliability, we manually verify the generated analysis texts and expand the dataset using an additional data engine.

From a model perspective, FakeShield employs a multimodal forgery localization module (MFLM) that leverages the segment anything model (SAM) and a large language model to localize tampered regions based on their general vision and language features. However, recent studies \cite{MVSS-Net, CAT-Net2, SAFL-Net} highlight the crucial role of low-level noise features in image manipulation detection, while pre-trained models for general vision tasks primarily capture high-level semantic features. To address this challenge, we introduce a trace encoder with noise enhancement to compensate for the limitations of the base vision model. In addition, we design a dedicated fusion mechanism to better integrate multiple features. Ablation studies in Table \ref{table:exp_localization_ablation} further validate the effectiveness of these components.

In the following experiments, we comprehensively compare FakeShield and our method, ForgerySleuth, demonstrating that ForgerySleuth consistently achieves superior performance in various evaluation aspects.

\subsection{Evaluation}

\vspace{0.05in}
\noindent\textbf{Experimental Setting.} We reproduce the FakeShield model using its publicly released official code repository and pre-trained weights. The following comparative experiments adopt the same experimental settings and evaluation metrics as described in Section \ref{sec:experiment_setting}.

We follow established practices \cite{UnionFormer} by using optimal threshold and fixed threshold F1 scores and the threshold-independent Area Under the Curve (AUC) metric. Inspired by prior work \cite{Vicuna, LLaVA}, we use GPT-4 as an automated evaluator to assess the reasoning performance of different models. To address concerns regarding potential limitations of GPT-4, we incorporate an additional metric, semantic textual similarity (STS). Specifically, following SBERT \cite{SBERT}, we use STS to measure the similarity between the generated analysis and the ground-truth text in ForgeryAnalysis-Eval.

\begin{table}[t]
    \centering
    \caption{Forgery analysis results of ForgerySleuth and SoTA methods, evaluated using GPT-4 on ForgeryAnalysis-Eval dataset.}
    \vspace{0.5em}
    \resizebox{0.6\linewidth}{!}{
        \setlength{\tabcolsep}{1.5mm}{
            \begin{tabular}{lccccc}
                \toprule
                \textbf{Method} & \textit{Recall} & Correctness & Relevance & Detail & Overall \\
                \midrule
                LISA \cite{LISA} & - & 2.27 & 2.51 & 1.12 & 1.92 \\
                QWen & 43.9\% & 3.22 & 2.92 & 3.10 & 3.41\\
                QWen2 \cite{Qwen2-VL} & 29.6\% & 3.70 & 4.14 & 2.57 & 3.34 \\
                GPT-4o & 68.8\% & 7.14 & 7.97 & 5.84 & 6.96 \\
                \rowcolor{gray!15}
                FakeShield \cite{FakeShield} & 52.8\% & 5.91 & 6.35 & 4.67 & 5.55 \\
                \rowcolor{gray!30}
                \textbf{ForgerySleuth} & \textbf{92.2\%} & \textbf{9.10} & \textbf{9.60} & \textbf{9.89} & \textbf{9.45} \\
                \bottomrule
            \end{tabular}
        }
    }
    \label{table_supp:exp_analysis_gpt4}
\end{table}

\vspace{0.05in}
\noindent\textbf{Manipulation Detection Results.} Tables \ref{table_supp:exp_localization_pretraining_F1} and \ref{table_supp:exp_localization_pretraining_AUC} present the performance of ForgerySleuth, FakeShield, and other SoTA methods in the manipulation localization task, reporting F1 scores and AUC metrics, respectively. It is essential to note that the official implementation of FakeShield follows a two-stage pipeline. First, it determines whether an image has been manipulated. Only if an image is classified as fake does it proceed to generate a tampering mask. In contrast, ForgerySleuth and other detection methods output a mask for all images while also providing a real/fake classification result.

We adopt the following strategy for FakeShield to maintain consistency with other methods and ensure a fair comparison. If FakeShield classifies an image as real, we assume that the predicted mask is an all-zero matrix (\textit{i.e.}, indicating no tampered region). The corresponding results are shown in the FakeShield row. The notably lower accuracy stems from the low recall rates, where FakeShield fails to detect manipulations, leading to significant missed detections. To further analyze FakeShield's localization performance, we also report its performance on only the successfully detected manipulated images (\textit{i.e.}, True Positive samples). As indicated in the row for only true positive samples, its performance improves significantly under this setting, but it still falls short compared to ForgerySleuth. However, since manipulation detection is performed without prior knowledge of whether an image is manipulated, the first setting (which considers all images) provides a more realistic and fair evaluation, while the second setting serves only as a supplementary analysis.

\begin{table}[t]
    \centering
    \caption{Forgery analysis results of ForgerySleuth and SoTA methods, evaluated using STS.}
    \vspace{0.5em}
    \resizebox{0.65\linewidth}{!}{
        \setlength{\tabcolsep}{1.4mm}{
            \begin{tabular}{l|cccccc}
                \toprule
                \textbf{Model} & \cellcolor{gray!30} \textbf{Ours} & \cellcolor{gray!15} FakeShield & GPT-4o & QWen2 & QWen & LISA \\
                \midrule
                all-MiniLM-L6-v2 & \cellcolor{gray!30} \textbf{0.926} & \cellcolor{gray!15} 0.662 & 0.725 & 0.595 & 0.551 & 0.313 \\
                all-MiniLM-L12-v2 & \cellcolor{gray!30} \textbf{0.919} & \cellcolor{gray!15} 0.514 & 0.645 & 0.505 & 0.475 & 0.350 \\
                all-mpnet-base-v2 & \cellcolor{gray!30} \textbf{0.961} & \cellcolor{gray!15} 0.695 & 0.724 & 0.635 & 0.546 & 0.401 \\
                \bottomrule
            \end{tabular}
        }
    }
    \label{table_supp:exp_analysis_sts}
\end{table}

\vspace{0.05in}
\noindent\textbf{Forgery Analysis Results.} We employ GPT-4 as an evaluator on the ForgeryAnalysis-Eval dataset to assess the quality of textual analysis and reasoning. The evaluation considers several key dimensions: correctness, relevance,
and detail. The results are presented in Table \ref{table_supp:exp_analysis_gpt4} (corresponding to Figure \ref{figure:teaser} (b)). FakeShield exhibits low recall rates, further corroborating the previously discussed issue of missed detections. Additionally, its analysis falls behind ForgerySleuth in both accuracy and level of detail.

However, since GPT-4 itself is susceptible to hallucinations, this approach does not guarantee a fully objective and quantitative evaluation. To address this limitation, we incorporate Semantic Textual Similarity (STS) to measure the similarity between the generated text and the ground-truth text in ForgeryAnalysis-Eval. The results, shown in Table \ref{table_supp:exp_analysis_sts}, further confirm that our ForgerySleuth consistently outperforms FakeShield.

\begin{figure*}[b]
  \centering
  \includegraphics[width=\linewidth]{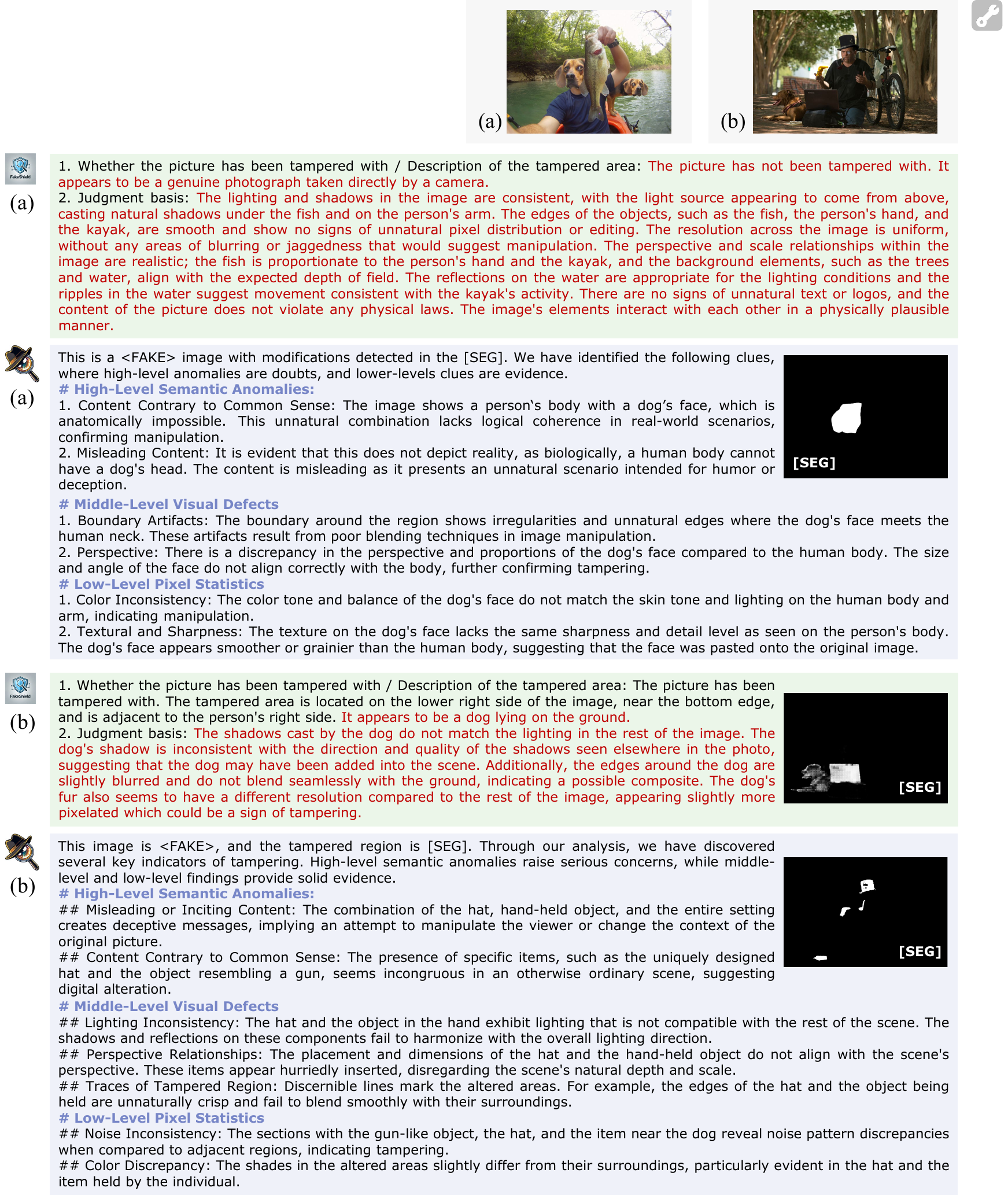}
  \caption{Comparison of results between ForgerySleuth and FakeShield.}
  \label{figure:sup_comparison_fakeshield}
\end{figure*}

\vspace{0.05in}
\noindent\textbf{Qualitative Results.} We present the performance of ForgerySleuth and FakeShield in manipulation localization and forgery analysis, as shown in Figure \ref{figure:sup_comparison_fakeshield}. The results demonstrate that ForgerySleuth not only achieves higher accuracy in localizing manipulated regions, but also generates more detailed and precise analysis.

\section{Experiment}

\subsection{Experimental Implementation Details}

\noindent\textbf{Implementation Details.} We employ LLaVA-7B-v1-1 \cite{LLaVA} as the base multimodal LLM ($\mathcal{F}_{m}$) and use the ViT-H SAM \cite{SAM} backbone for the vision encoder ($\mathcal{F}_{v}$). For training, we utilize 2 NVIDIA 80GB A800 GPUs, with training scripts optimized by DeepSpeed \cite{DeepSpeed}, which helps reduce memory usage and accelerate training. We use the AdamW \cite{AdamW} optimizer, setting the learning rate to $0.0002$ with no weight decay. The learning rate is scheduled using WarmupDecayLR, with 100 warmup iterations. The weights for the text generation loss $\lambda_\text{txt}$ and mask loss $\lambda_\text{mask}$ are both set to $1.0$, while the BCE loss $\lambda_\text{bce}$ and DICE loss $\lambda_\text{dice}$ are weighted at $1.0$ and $0.2$, respectively. The batch size per device is $4$, with gradient accumulation steps set to $4$.

\subsection{Manipulation Detection Results}

\begin{figure}[t]
  \centering
  \includegraphics[width=0.56\linewidth]{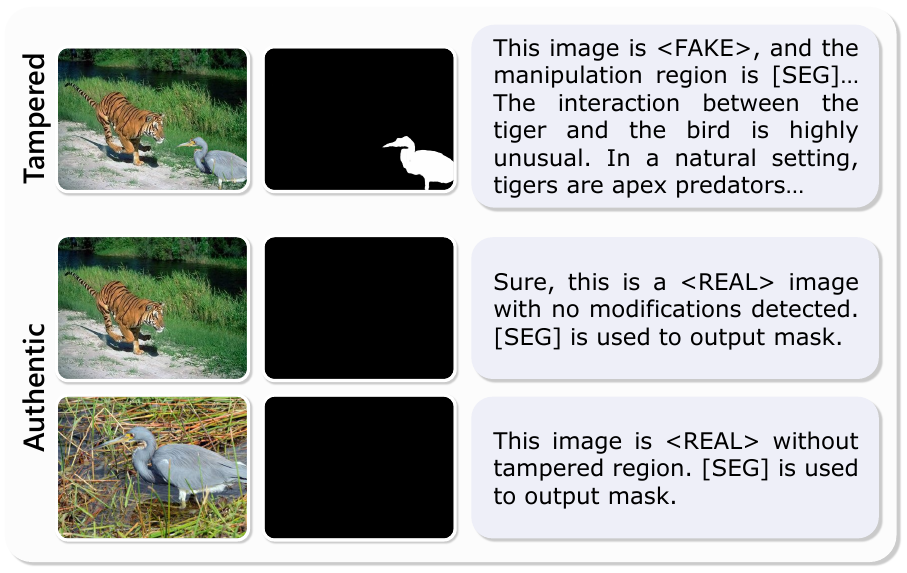}
  \caption{Predictions of ForgerySleuth on authentic and tampered images.}
  \label{figure:real_sample}
\end{figure}

Modeling the natural laws of real samples and distinguishing them from tampered images is crucial. We utilized two special tokens, \texttt{<REAL>} and \texttt{<FAKE>}, which indicate the detection results. In addition, the output mask provides valuable information for verifying image authenticity. Figure \ref{figure:real_sample} illustrates the ForgerySleuth detection results for authentic and manipulated images, including the predicted tampering masks and textual analysis. Our method achieves real/fake classification accuracies of 0.989 on Columbia and 0.910 on CASIA1, demonstrating its effectiveness in manipulation detection.

\subsection{Forgery Analysis Results}

\label{sec:appendices_evaluation}

\begin{figure*}[tbh]
  \centering
  \includegraphics[width=\linewidth]{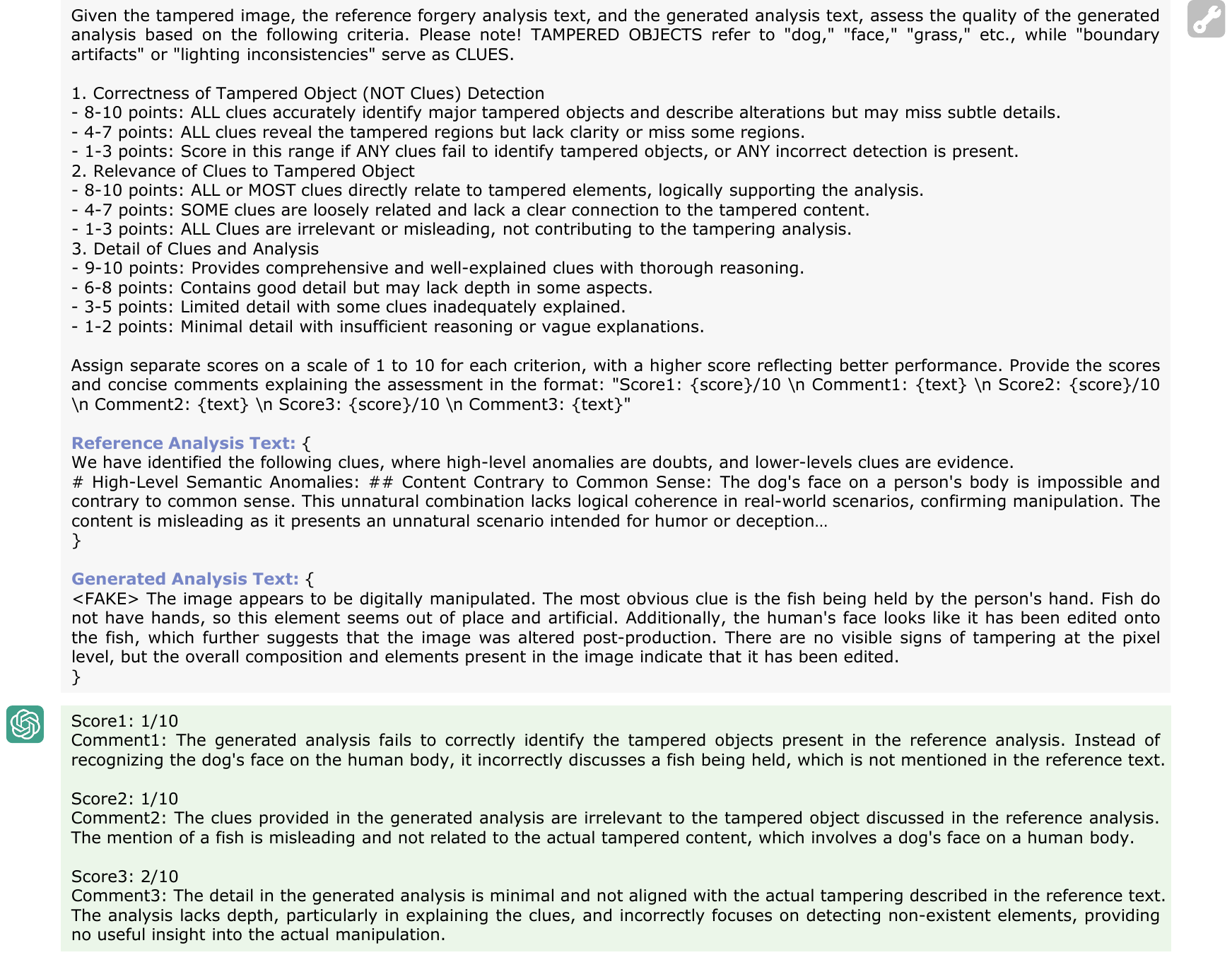}
  \caption{Illustration of the prompt and GPT-4 response of text analysis quality evaluation. The evaluation prompt includes clear scoring criteria for the assessment aspects, ensuring consistent and fair scoring.}
  \label{figure:sup_evaluation_gpt4}
\end{figure*}

We leverage GPT-4 for evaluation based on the ForgeryAnalysis-Eval dataset to assess the quality of text reasoning and explanations. Ratings are based on several dimensions, including the correctness of tampered objects, the relevance of clues to manipulation, and the detail of analysis, reflecting the capability of comprehension, reasoning, and correctness. We collect responses from each M-LLM, and GPT-4 assigns scores from 1 to 10 for each response, with higher scores indicating better performance. Beyond scoring, GPT-4 provides explanations for its ratings, ensuring transparency and consistency in the evaluation process. Figure \ref{figure:sup_evaluation_gpt4} illustrates the prompt structure used for the evaluation and the response of GPT-4. To ensure consistent and fair scoring, the evaluation prompt includes clear scoring criteria for assessment aspects, including correctness, relevance, and details. GPT-4 assigns a score for each evaluation dimension and provides detailed comments to justify the rating.

To directly assess the ability to detect manipulated images, we explicitly instruct the evaluated M-LLMs through prompts to additionally output \texttt{<FAKE>} or \texttt{<REAL>} to indicate their detection results. All models, except LISA-7B-v1-explanatory, can provide the required response. We use the recall rate to reflect the ability to correctly identify tampered images.

\vspace{0.05in}
\noindent\textbf{Human Evaluations.} We also conduct human evaluations under the same experimental setup as the evaluation using GPT-4 described above. A total of 14 volunteers participated, each scoring for 30 random samples. The results are summarized in Table \ref{table_supp:exp_analysis_user_study}.

\begin{table}[t]
    \centering
    \caption{Human evaluation results for the forgery analysis task.}
    \vspace{0.5em}
    \resizebox{0.6\linewidth}{!}{
        \setlength{\tabcolsep}{1.2mm}{
            \begin{tabular}{lcccccc}
                \toprule
                \textbf{Method} & Samples & \textit{Recall} & Correctness & Relevance & Detail & Overall \\
                \midrule
                QWen2 & 139 & 35.3\% & 4.95 & 5.10 & 2.85 & 4.11 \\
                GPT-4o & 134 & 65.7\% & 8.06 & 8.32 & 5.77 & 7.18 \\
                \textbf{Ours} & 147 & \textbf{91.2\%} & \textbf{9.48} & \textbf{9.53} & \textbf{9.61} & \textbf{9.44} \\
                \bottomrule
            \end{tabular}
        }
    }
    \label{table_supp:exp_analysis_user_study}
\end{table}

\subsection{Qualitative Results}

We present additional results of our ForgerySleuth in both forgery analysis and manipulation localization tasks. We also provide forgery analysis results from existing M-LLMs (GPT-4o and Qwen-VL) and segmentation masks from traditional image manipulation detection (IMD) methods, illustrating the advantages of our proposed IMD assistant in terms of accuracy and explainability. These cases are from public datasets IMD20 and NIST16, demonstrating the generalization capabilities of our method.

\begin{figure*}[t]
  \centering
  \includegraphics[width=\linewidth]{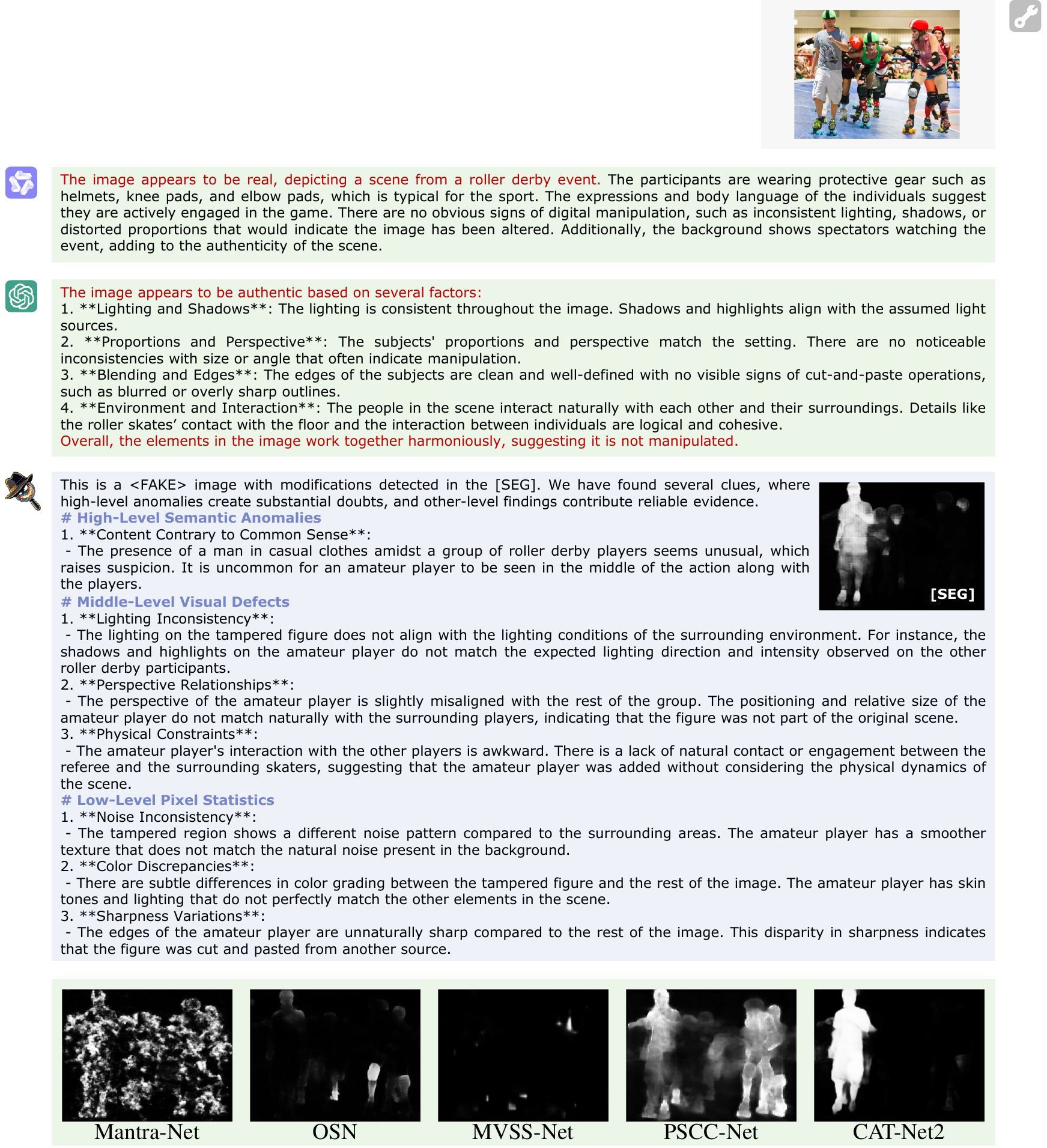}
  \caption{Comparison of results between our ForgerySleuth, existing multimodal large language models, and traditional image manipulation detection methods. The image is c93gg4y\_0.png from the IMD20 dataset.}
  \label{figure:sup_pred2}
\end{figure*}

\begin{figure*}[t]
  \centering
  \includegraphics[width=\linewidth]{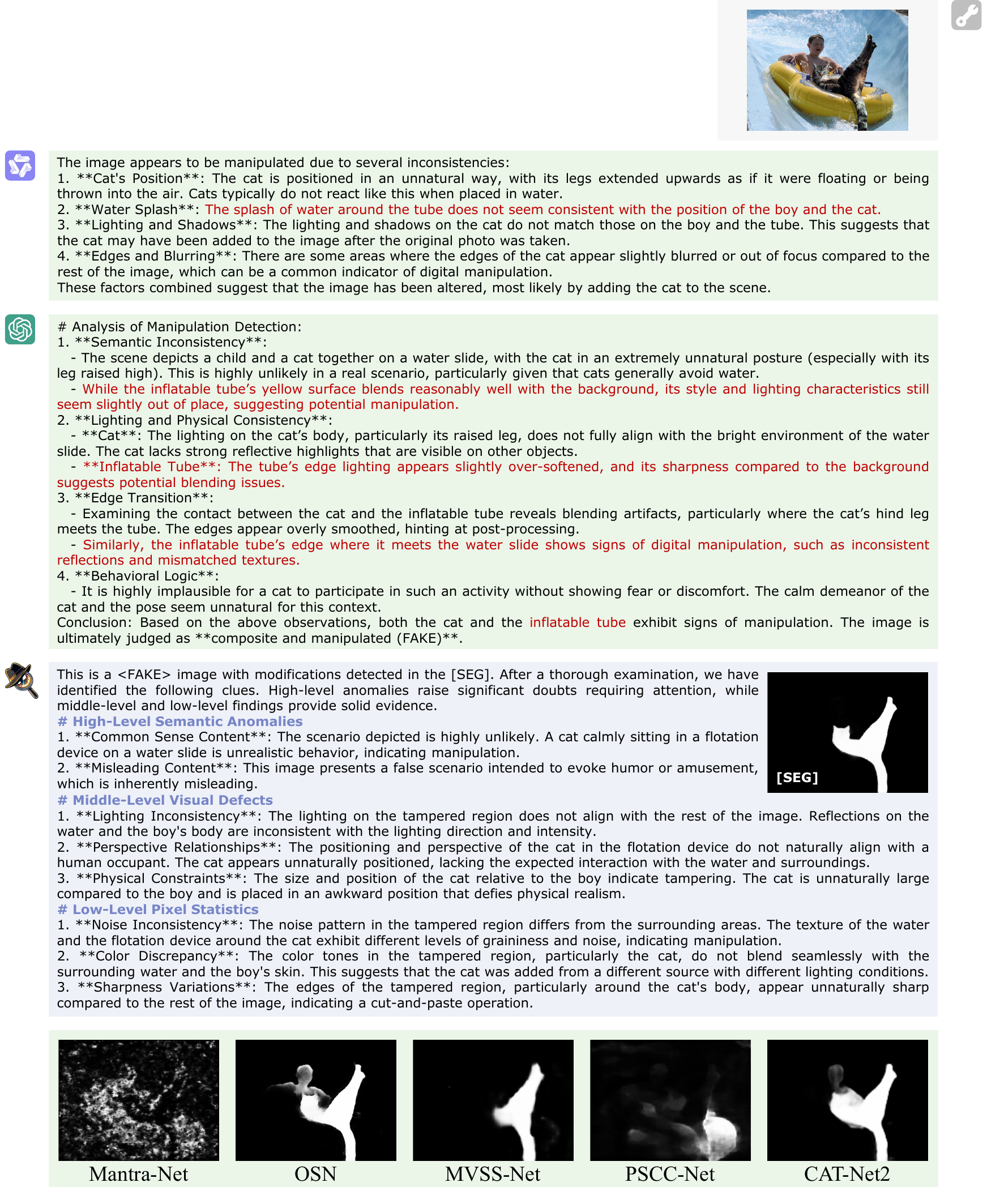}
  \caption{Comparison of results between our ForgerySleuth, existing multimodal large language models, and traditional image manipulation detection methods. The image is c8zf0hg\_0.jpg from the IMD20 dataset.}
  \label{figure:sup_pred1}
\end{figure*}

\begin{figure*}[t]
  \centering
  \includegraphics[width=\linewidth]{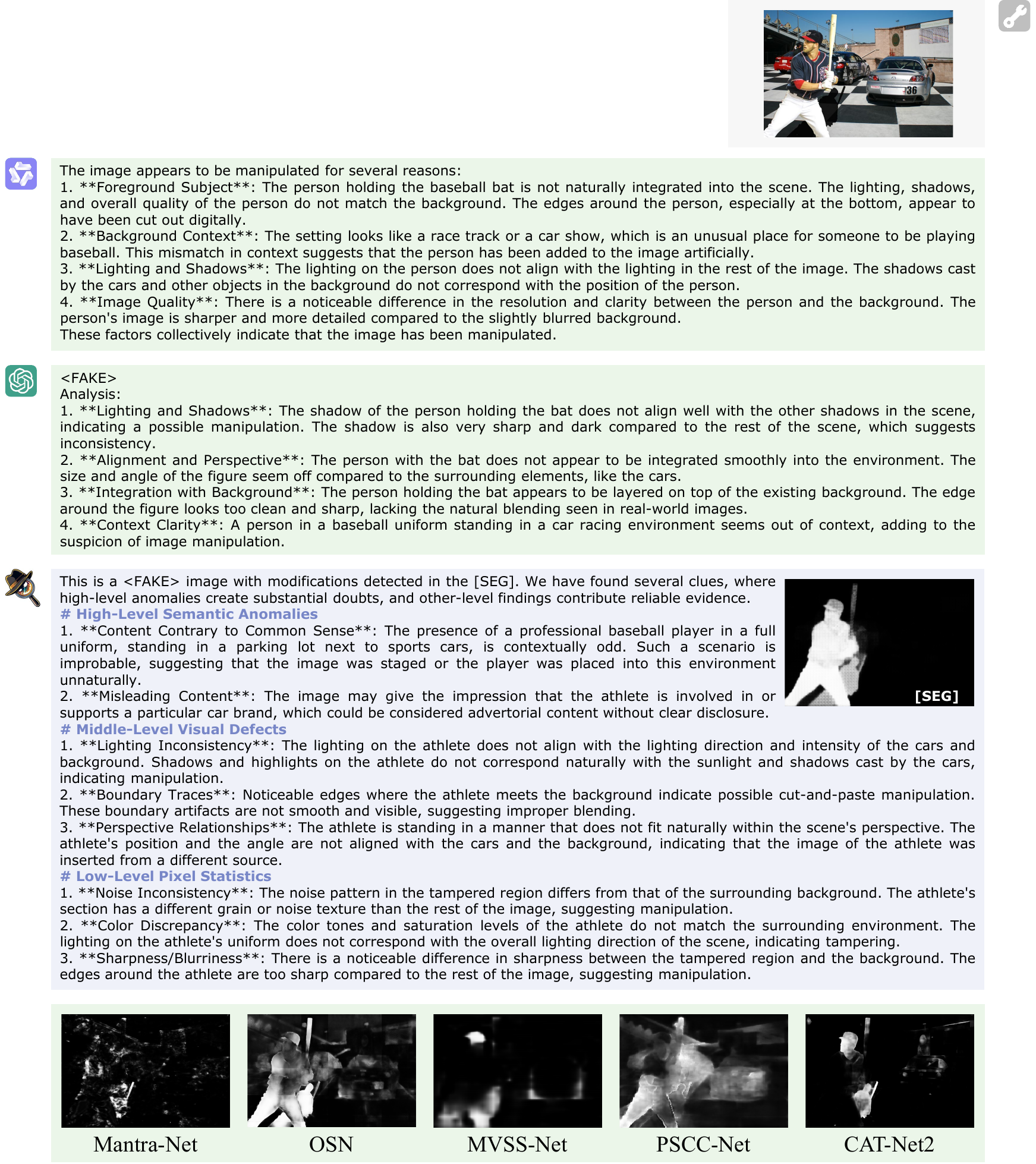}
  \caption{Comparison of results between our ForgerySleuth, existing multimodal large language models, and traditional image manipulation detection methods. The image is NC2016\_7312.jpg from the NIST16 dataset.}
  \label{figure:sup_pred3}
\end{figure*}

In the example in Figure \ref{figure:sup_pred2}, both M-LLMs classify the image as real without detecting manipulation. Similarly, most IMD methods, except CAT-Net2, fail to localize the manipulated regions accurately. However, ForgerySleuth identifies the tampered regions and provided a detailed tampering analysis. Figure \ref{figure:sup_pred1} presents a case with more apparent tampering traces. GPT-4o and Qwen-VL both exhibit varying degrees of overthinking, leading to inaccurate analysis. Our method demonstrates higher localization precision and analytical accuracy. In the case of Figure \ref{figure:sup_pred3}, ForgerySleuth precisely localizes the tampered regions, even including the shadow of the person. Compared to the textual outputs of M-LLMs, the mask generated by ForgerySleuth provides a more intuitive and accurate detection result.

\section{Limitation and Future Work} 
\label{sec:limitation}

\subsection{Failure Cases} 

\begin{figure}[t]
  \centering
  \includegraphics[width=0.6\linewidth]{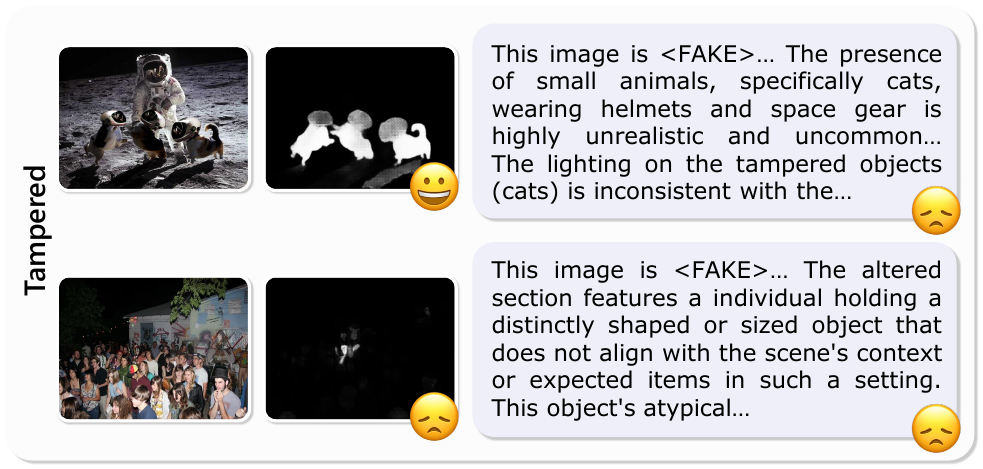}
  \caption{Failure cases in predictions of ForgerySleuth.}
  \label{figure:failure_cases}
\end{figure}

We analyzed ForgerySleuth's detection failures and identified two representative error patterns.

\vspace{0.05in}
\noindent\textbf{Semantic Description Inaccuracies.} In these cases, our model can accurately localize manipulated regions, but its textual analysis contains description errors. For instance, in the first example in Figure \ref{figure:failure_cases}, the model misidentifies dogs as cats. Importantly, this does not impact the detection logic or localization accuracy. These errors often arise when specific semantic concepts in images are challenging to detect clearly. Future work could improve this by enhancing the visual understanding capabilities of MLLM and LLM backbone models.

\vspace{0.05in}
\noindent\textbf{Failures with Small Manipulations in Complex Scenes.} In complex image content with small manipulated regions, the model's localization precision can degrade, sometimes leading to incorrect textual analysis and, consequently, a detection failure. For example, in the second instance in Figure \ref{figure:failure_cases}, where a spliced cartoon character is present, the model only localized a partial manipulation (\textit{e.g.}, the guitar in hand) while overlooking the character itself, primarily due to the small proportion of the manipulation and the image's overall complexity. Detecting small region manipulations remains a significant challenge, and MLLMs tend to focus on global semantic information rather than fine-grained local details. Therefore, developing methods to effectively detect anomalies in small regions is a crucial research direction.

\subsection{Constraints in Evaluation Methods} 

Our paper evaluates analysis accuracy using GPT-4o and STS methods on the ForgeryAnalysis-eval dataset. However, our evaluation methods are subject to specific constraints: our assessment relies on GPT-4o, a large external model that may exhibit inherent biases or hallucinations, and STS only reflects semantic similarity between two sentences. Thus, exploring more comprehensive and objective evaluation methods, including the quantitative assessment of model hallucinations, is an important direction for future work.

\subsection{Model Scale and Inference Costs} 

While leveraging LLMs to complement traditional image manipulation detection methods, which often primarily focus on low-level traces, this approach increases the model's scale and computational cost of generating detailed explanations. In the future, we plan to investigate model light-weighting solutions. This includes developing a version where generating such detailed output is optional, which would significantly reduce inference time. Alternatively, we could explore distilling LLM capabilities into smaller models, retaining only the essential knowledge for manipulation detection.

\newpage
\clearpage
\section*{NeurIPS Paper Checklist}

\begin{enumerate}

\item {\bf Claims}
    \item[] Question: Do the main claims made in the abstract and introduction accurately reflect the paper's contributions and scope?
    \item[] Answer: \answerYes{} % Replace by \answerYes{}, \answerNo{}, or \answerNA{}.
    \item[] Justification: The abstract and introduction accurately state our main contributions, the scope of our method, and these are well-supported by the results presented in the paper.

\item {\bf Limitations}
    \item[] Question: Does the paper discuss the limitations of the work performed by the authors?
    \item[] Answer: \answerYes{} % Replace by \answerYes{}, \answerNo{}, or \answerNA{}.
    \item[] Justification: We discuss the limitations of our work in the \textit{Appendices} due to page constraints.

\item {\bf Theory assumptions and proofs}
    \item[] Question: For each theoretical result, does the paper provide the full set of assumptions and a complete (and correct) proof?
    \item[] Answer: \answerNA{} % Replace by \answerYes{}, \answerNo{}, or \answerNA{}.
    \item[] Justification: This paper focuses on a novel method and a new dataset for image manipulation detection and does not include theoretical results like theorems or formal proofs.

\item {\bf Experimental result reproducibility}
    \item[] Question: Does the paper fully disclose all the information needed to reproduce the main experimental results of the paper to the extent that it affects the main claims and/or conclusions of the paper (regardless of whether the code and data are provided or not)?
    \item[] Answer: \answerYes{} % Replace by \answerYes{}, \answerNo{}, or \answerNA{}.
    \item[] Justification: We provide detailed descriptions of our method, relevant training configurations, and evaluation protocols. For our novel dataset, we thoroughly describe its construction, composition, and splits. These details are sufficient for reproducing our main experimental results. 

\item {\bf Open access to data and code}
    \item[] Question: Does the paper provide open access to the data and code, with sufficient instructions to faithfully reproduce the main experimental results, as described in supplemental material?
    \item[] Answer: \answerYes{} % Replace by \answerYes{}, \answerNo{}, or \answerNA{}.
    \item[] Justification: Our code, data, and model weights are already open-sourced. To facilitate reproducibility while ensuring reviewer anonymity, we provide access to the code and data as described in our supplementary material. While the pre-trained model weights are also open-sourced, due to their significant size, they are not directly included in the current supplementary package. We are committed to providing public links for resources in the abstract of the camera-ready paper.

\item {\bf Experimental setting/details}
    \item[] Question: Does the paper specify all the training and test details (e.g., data splits, hyperparameters, how they were chosen, type of optimizer, etc.) necessary to understand the results?
    \item[] Answer: \answerYes{} % Replace by \answerYes{}, \answerNo{}, or \answerNA{}.
    \item[] Justification: Comprehensive details, including hyperparameters, optimizer types, learning rates, batch sizes, and other critical training and testing specifics, are thoroughly introduced in the \textit{Appendices}.

\item {\bf Experiment statistical significance}
    \item[] Question: Does the paper report error bars suitably and correctly defined or other appropriate information about the statistical significance of the experiments?
    \item[] Answer: \answerYes{} % Replace by \answerYes{}, \answerNo{}, or \answerNA{}.
    \item[] Justification: To ensure fair comparisons, our experimental setup and evaluation protocols are consistent with established state-of-the-art methods in the field. While comprehensive statistical significance tests are not uniformly standard practice in this research domain, we address potential experimental variability. For experiments involving stochastic elements, we conduct evaluations twice and report the average performance.

\item {\bf Experiments compute resources}
    \item[] Question: For each experiment, does the paper provide sufficient information on the computer resources (type of compute workers, memory, time of execution) needed to reproduce the experiments?
    \item[] Answer: \answerYes{} % Replace by \answerYes{}, \answerNo{}, or \answerNA{}.
    \item[] Justification: We provide details on the computational resources used for our experiments.
    
\item {\bf Code of ethics}
    \item[] Question: Does the research conducted in the paper conform, in every respect, with the NeurIPS Code of Ethics \url{https://neurips.cc/public/EthicsGuidelines}?
    \item[] Answer: \answerYes{} % Replace by \answerYes{}, \answerNo{}, or \answerNA{}.
    \item[] Justification: The research presented in this paper has been conducted with careful consideration of the NeurIPS Code of Ethics.

\item {\bf Broader impacts}
    \item[] Question: Does the paper discuss both potential positive societal impacts and negative societal impacts of the work performed?
    \item[] Answer: \answerYes{} % Replace by \answerYes{}, \answerNo{}, or \answerNA{}.
    \item[] Justification: This paper discusses the broader societal impacts of our work. A significant positive impact is the potential to enhance the authenticity of visual content in media, which can help combat the spread of misinformation. However, we also acknowledge potential negative impacts. One key concern is the possibility of inaccurate detections, which could lead to the mislabeling of legitimate content.
    
\item {\bf Safeguards}
    \item[] Question: Does the paper describe safeguards that have been put in place for responsible release of data or models that have a high risk for misuse (e.g., pretrained language models, image generators, or scraped datasets)?
    \item[] Answer: \answerYes{} % Replace by \answerYes{}, \answerNo{}, or \answerNA{}.
    \item[] Justification: We have implemented several safeguards. The model weights are hosted on an authoritative platform. Moreover, access to the data is controlled via an application process, which allows us to carefully review and approve user requests.

\item {\bf Licenses for existing assets}
    \item[] Question: Are the creators or original owners of assets (e.g., code, data, models), used in the paper, properly credited and are the license and terms of use explicitly mentioned and properly respected?
    \item[] Answer: \answerYes{} % Replace by \answerYes{}, \answerNo{}, or \answerNA{}.
    \item[] Justification: For every asset used in this paper, we have properly cited the corresponding original paper or source. We have also ensured that our use fully complies with the respective licenses and terms of use.

\item {\bf New assets}
    \item[] Question: Are new assets introduced in the paper well documented and is the documentation provided alongside the assets?
    \item[] Answer: \answerYes{} % Replace by \answerYes{}, \answerNo{}, or \answerNA{}.
    \item[] Justification: All assets introduced in this paper are well documented, and this documentation is provided alongside the assets.

\item {\bf Crowdsourcing and research with human subjects}
    \item[] Question: For crowdsourcing experiments and research with human subjects, does the paper include the full text of instructions given to participants and screenshots, if applicable, as well as details about compensation (if any)? 
    \item[] Answer: \answerYes{} % Replace by \answerYes{}, \answerNo{}, or \answerNA{}.
    \item[] Justification: The instructions given to participants are provided in the \textit{appendices}.

\item {\bf Institutional review board (IRB) approvals or equivalent for research with human subjects}
    \item[] Question: Does the paper describe potential risks incurred by study participants, whether such risks were disclosed to the subjects, and whether Institutional Review Board (IRB) approvals (or an equivalent approval/review based on the requirements of your country or institution) were obtained?
    \item[] Answer: \answerYes{} % Replace by \answerYes{}, \answerNo{}, or \answerNA{}.
    \item[] Justification: The potential risks involved in our user study, such as time expenditure, were disclosed to the participants within the annotation system and user experiment system. The experiments, conducted under limited risk conditions, received approval from our laboratory's review process.

\item {\bf Declaration of LLM usage}
    \item[] Question: Does the paper describe the usage of LLMs if it is an important, original, or non-standard component of the core methods in this research? Note that if the LLM is used only for writing, editing, or formatting purposes and does not impact the core methodology, scientific rigorousness, or originality of the research, declaration is not required.
    %this research? 
    \item[] Answer: \answerYes{} % Replace by \answerYes{}, \answerNo{}, or \answerNA{}.
    \item[] Justification: Large Language Models (LLMs) constitute an important component of the core methodology in this research. The relevant aspects of their usage are described in detail within the main body of the paper.

\end{enumerate}

\end{document}